\documentclass[conference]{IEEEtran}
\IEEEoverridecommandlockouts

\usepackage{adjustbox}
\usepackage{amsmath,amssymb,amsfonts}
\usepackage[linesnumbered,ruled,vlined]{algorithm2e}
\usepackage{setspace}
\usepackage{algorithmic}
\usepackage{float}
\usepackage{graphicx}
\usepackage{textcomp}
\usepackage{xcolor}
\usepackage{multirow}
\usepackage{mdframed}
\usepackage{balance}
\usepackage{booktabs}
\usepackage{tcolorbox}
\usepackage{makecell}
\usepackage{threeparttable}
\usepackage{colortbl}
\usepackage{subfigure}
\usepackage{hhline}
\usepackage{pifont}
\usepackage{listings}
\usepackage{enumitem}

\usepackage{url}

\newcommand{\tabincell}[2]{\begin{tabular}{@{}#1@{}}#2\end{tabular}}

\newcommand{\parabf}[1]{\noindent\textbf{#1}}

\definecolor{ggray}{HTML}{eff0f0}
\definecolor{gggray}{HTML}{E8E8E8}
\definecolor{ggggray}{HTML}{BEBEBE}

\newcommand{\et}{\textit{et al.}}
\newcommand{\ie}{\textit{i.e.,}}
\newcommand{\eg}{\textit{e.g.,}}

\newcommand{\llm}{LLM\xspace}

\newcommand{\vicuna}{Vicuna\xspace}
\newcommand{\alpaca}{Alpaca\xspace}
\newcommand{\codealpaca}{CodeAlpaca\xspace}
\newcommand{\dolly}{Dolly\xspace}
\newcommand{\stablelm}{StableLM\xspace}
\newcommand{\chatglm}{ChatGLM\xspace}
\newcommand{\codegen}{CodeGen\xspace}
\newcommand{\instructcodegen}{Instruct-CodeGen\xspace}

\newcommand{\preferrate}{PR\xspace}

\newcommand{\pythia}{Pythia\xspace}

\newcommand{\wizardcoder}{WizardCoder\xspace}
\newcommand{\llama}{LLaMA\xspace}
\newcommand{\lora}{LoRA\xspace}

\usepackage[utf8]{inputenc}
\usepackage[english]{babel}

\newcounter{finding}
\newcommand{\finding}[1]{\refstepcounter{finding}
 	\vspace{1mm}
	\begin{mdframed}[linecolor=gray,roundcorner=12pt,backgroundcolor=gray!15,linewidth=3pt,innerleftmargin=2pt, leftmargin=0cm,rightmargin=0cm,topline=false,bottomline=false,rightline = false]
		\textbf{Finding \arabic{finding}:} #1
	\end{mdframed}
	\vspace{1mm}
}

\usepackage{footmisc}

\newcommand{\distance}{5pt}
\begin{document}

\title{Evaluating Instruction-Tuned Large Language Models  on Code Comprehension and Generation} 


\author{ \IEEEauthorblockN{
Zhiqiang Yuan,
Junwei Liu, 
Qiancheng Zi, 
Mingwei Liu,
Xin Peng,
Yiling Lou}

\IEEEauthorblockA{Department of Computer Science, Fudan University, China}
\IEEEauthorblockA{\{zhiqiangyuan23, jwliu22, qczi22\}@m.fudan.edu.cn}
\IEEEauthorblockA{\{liumingwei, pengxin, yilinglou\}@fudan.edu.cn}
}

\maketitle

\begin{abstract}
Instruction tuning has been proposed to enhance the generalization capability of large language models (LLMs) on new downstream tasks. To date, many efforts have been dedicated into evaluating instructed LLMs, covering not only general NLP tasks but also specific domains. However, little evaluation of instructed LLMs is diving into the software engineering domain, except the NL-to-Code task (generating a function for the given natural language description), which is only one of the code-related tasks in software development and maintenance. Although some recent work explores the capability of the instructed models such as ChatGPT on SE tasks, these commercial models are closed-source, thus lacking transparency and reproducibility. Overall, it still remains unclear how the recent open-source instructed LLMs perform on diverse code comprehension and generation tasks. 

To fill this knowledge gap, we  evaluate 10 open-source instructed LLMs on four representative code comprehension and generation tasks (\ie{} defect detection, clone detection, assertion generation, and code summarization). We have the following main findings. First, for the \textit{zero-shot} setting,  instructed LLMs are very competitive on code comprehension and generation tasks and sometimes even better than small SOTA models specifically fine-tuned on each downstream task. We also find that on code-related tasks LLMs instructed by code domain do not necessarily outperform LLMs instructed by general domain, and larger instructed LLMs are not always better. Second, for the \textit{few-shot} setting, we find that adding demonstration examples substantially helps instructed LLMs perform better on most code comprehension and generation tasks; however, the examples would sometimes induce unstable or even worse performance. In addition, we observe a performance drop with the increasing input length and an increasing instruction-following capability in the few-shot setting. Furthermore, we find the widely-used BM25-based shot selection strategy significantly outperforms the basic random selection or fixed selection only on generation problems (\eg{} assertion generation and code summarization), while exhibiting no significant difference from either  basic strategies on classification problems (\eg{} defect detection or clone detection). Third, for the \textit{fine-tuning} setting, we find that fine-tuning could further improve the model performance on downstream code comprehension and generation tasks compared to the zero-shot/one-shot performance. In addition, after being fine-tuned on the same downstream task dataset, instructed LLMs outperform both the small SOTA models and similar-scaled LLMs without instruction tuning. Based on our findings, we further present practical implications on model and usage recommendation, performance and cost trade-offs, and future direction.

\end{abstract}

\vspace{-5mm}
\section{Introduction}

Large Language Models (LLMs) have achieved great advance and have been leveraged in various domains~\cite{brown2020language, touvron2023llama, workshop2023bloom, chowdhery2022palm}.
More recently, instruction tuning has been proposed to enhance the generalization capability of LLMs on new downstream tasks~\cite{vicuna2023, alpaca, codealpaca, sanh2022multitask, ouyang2022training}. In particular, instruction tuning fine-tunes the pre-trained LLMs (also known as foundation models) with massive instructions from multiple tasks, and the LLMs after instruction tuning (denoted as instruction-tuned LLMs or instructed LLMs for short) are able to solve various unseen tasks in the zero-shot scenario without further fine-tuning or demonstration examples~\cite{wei2022finetuned, ouyang2022training, sanh2022multitask}. For example, ChatGPT~\cite{chatgpt}, a popular instructed LLM, is closed-source and supported by OpenAI. In addition, there has been a feverish trend of developing  instructed LLMs in the open-source community, \eg{} researchers have proposed multiple open-source instructed LLMs (\eg{} \vicuna{}~\cite{vicuna2023}, \alpaca{}~\cite{alpaca}, and \codealpaca{}~\cite{codealpaca}) by fine-tuning the open-source foundation model \llama{}~\cite{touvron2023llama} with instruction datasets from general
or code-specific domains.

Amidst the rapid development and increasing popularity of instruction-tuned LLMs, it becomes imperative to grasp the capabilities of these novel models. To date, many research efforts have been put into evaluating instructed LLMs~\cite{DBLP:journals/corr/abs-2307-03109, DBLP:journals/corr/abs-2306-04757,zhao2023survey, ji2023exploring, chung2022scaling}, including not only general NLP tasks~\cite{raheja2023coedit,chakrabarty2022help} (\eg{} sentiment analysis, text classification, and semantic understanding) but also specific domains~\cite{labrak2023zeroshot} (\eg{} medical, education, and agent applications). However, little evaluation of instructed LLMs is diving into the software engineering domain. Most evaluation only focuses on the coding capability of instructed LLMs~\cite{chen2021evaluating,50670,touvron2023llama, luo2023wizardcoder} by evaluating the model capability of generating a code snippet for a given natural language description, which is only one of the diverse code-related tasks in software development and maintenance. Although some recent work explores the capability of the instructed models such as ChatGPT and Codex~\cite{DBLP:conf/icse/LemieuxILS23, DBLP:journals/corr/abs-2307-04346, DBLP:conf/icse/KangYY23,zan2023large} on software engineering tasks (\eg{} testing and debugging), these commercial models are closed-source, thus lacking transparency and reproducibility. Overall, it still remains unclear how the recent open-source instructed LLMs perform on diverse code comprehension and generation tasks in software engineering.

To fill this knowledge gap, we make the first attempt to evaluate open-source instruction-tuned LLMs on code comprehension and generation tasks. 
While there has always been researched enthusiasm for leveraging advanced deep learning techniques to solve code-related tasks in software engineering domain~\cite{DBLP:conf/icse/TufanoPB23,zeng2022extensive,niu2023empirical, DBLP:conf/sigsoft/WangYGP0L22}, existing work focuses on relatively-small pre-trained models without instruction tuning (\eg{} CodeT5~\cite{wang2021codet5} and CodeBERT~\cite{feng2020codebert}). 
Different from existing work, we concentrate on recent instruction-tuned models, shedding light on the capability of instruction tuning in both code comprehension and generation.

In this work, we perform a comprehensive study for 10 state-of-the-art instruction-tuned LLMs on 4 representative code comprehension and generation tasks, \ie{} defect detection, clone detection, assertion generation, and code summarization. In particular, our studied instructed LLMs include the open-source ones newly-released in the past four months (\ie{} March 1st to July 1st 2023), covering a wide range of models scales (\ie{} from 6B to 16B parameters) based on different foundation models (\eg{} \llama{}~\cite{touvron2023llama}, \pythia{}~\cite{biderman2023pythia}, and GLM~\cite{zeng2022glm130b}) tuned by instructions from both general and code domains. In addition, we further include four small pre-trained models (\ie{} CodeGPT-adapted~\cite{codegpt}, CoText~\cite{phan2021cotext}, PLBART~\cite{ahmad-etal-2021-unified}, and CodeT5~\cite{wang2021codet5}) that have been shown to achieve the best performance after being fine-tuned on each studied task~\cite{niu2023empirical} and an additional large model without instruction tuning (\ie{} \codegen{}-6B~\cite{nijkamp2022codegen}) as baselines, so as to compare the performance of instructed models with SOTA models. For all the studied instruction-tuned LLMs, we evaluate their capability in three different settings by answering the following research questions.

\vspace{-2mm}
\begin{itemize}[leftmargin=12pt, topsep=5pt]
    \item \textbf{RQ1: How do instruction-tuned LLMs perform on code comprehension and generation tasks in the \textit{zero-shot} setting?} This RQ aims at investigating the zero-shot generalization of instructed LLMs on code-related tasks. 

    \item \textbf{RQ2: How do instruction-tuned LLMs perform on code comprehension and generation tasks in the \textit{few-shot} setting?} This RQ aims at investigating the in-context learning capability of instruction-tuned LLMs on code-related tasks when additional demonstration examples are given in the prompt. In addition, we also study the impact of three different shot-selection strategies. 

    \item \textbf{RQ3: How do instruction-tuned LLMs perform on code comprehension and generation tasks with further fine-tuning?} This RQ aims at exploring the performance of instructed LLMs after they are further fine-tuned on each specific code-related task. 
\end{itemize}

In addition, we also investigate the memory costs and time costs of using these instructed models for the community's reference via the following research question. 

\begin{itemize}[leftmargin=10pt, topsep=5pt]
    \item\textbf{RQ4: How are the costs of instruction-tuned LLMs during fine-tuning and inference?} 
\end{itemize}


\textbf{\textit{Main findings and implications.}} Based on our results, we have the following main findings. First, for the \textit{zero-shot} setting, we find that instructed LLMs are very competitive on code comprehension and generation tasks and sometimes even better than small SOTA models specifically fine-tuned on each downstream task. We also have interesting findings that on code-related tasks LLMs instructed by code domain do not necessarily outperform LLMs instructed by general domain, and larger instructed LLMs are not always better. Second, for the \textit{few-shot} setting, we find that adding demonstration examples substantially helps instructed LLMs perform better on most code comprehension and generation tasks; however, the examples would sometimes induce unstable or even worse performance. In addition, we observe a performance drop with the increasing input length and an increasing instruction-following capability in the few-shot setting. Furthermore, we find the widely-used BM25-based shot selection strategy significantly outperforms the basic random selection or fixed selection only on generation problems (\eg{} assertion generation and code summarization), while exhibiting no significant difference from either  basic strategies on classification problems (\eg{} defect detection or clone detection). Third, for the \textit{fine-tuning} setting, we find that fine-tuning could further improve the model performance on downstream code comprehension and generation tasks compared to the zero-shot/one-shot performance. In addition, after being fine-tuned on the same downstream task dataset, instructed LLMs outperform both the small SOTA models and similar-scaled LLMs without instruction tuning, suggesting the large benefits of the instruction tuning. Lastly, we find that similar-scaled instruction-tuned LLMs vary in memory costs and time costs; while they do not necessarily take more memory resources than small SOTA models, but take much more time costs than small SOTA models in both fine-tuning and inference. Based on our findings, we further present practical implications on model and usage recommendation, performance and cost trade-offs, and future directions. 

In summary, this work makes the following contributions: 
\begin{itemize}[leftmargin=12pt, topsep=5pt]
    \item To the best of our knowledge, this paper serves as the first study on evaluating \textit{open-source instruction-tuned LLMs} on code comprehension and generation tasks. In particular, we perform a large-scale experiment of 10 instructed LLMs with five baseline models on four representation code-related tasks in zero-shot, few-shot, and fine-tuning settings, which \textit{takes around 1000 GPU-hours on one NVIDIA A800-80GB GPU.}
    
    \item Our study shows many findings and practical implications on instructed LLMs for code comprehension and generation, such as the comparison between instructed LLMs and SOTA models in different settings, recommendations for instructed \llm{}s and shot select strategies, and trade-offs between model performance and costs.
    
\end{itemize}

\section{background}
\begin{figure}[htb]
    \centering
    \includegraphics[width=0.47\textwidth]{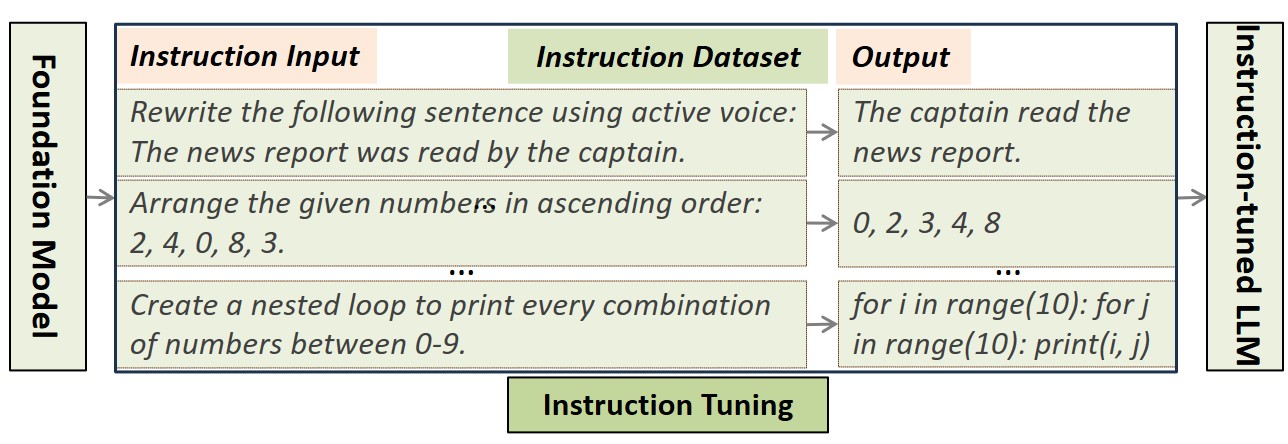}
    \caption{Instruction tuning}
    \label{figure:instruction_tuning}
\end{figure}

\textbf{\textit{Instruction tuning.}} Large language models (LLMs) are models with hundreds of billions (or more) of parameters, pre-trained on vast text datasets~\cite{wei2023chainofthought, shanahan2023talking, hoffmann2022training, taylor2022galactica}. 
To enhance their generalization ability to unseen tasks, LLMs serve as the foundation models for further tuning with the instruction dataset, a process known as instruction tuning~\cite{wei2022finetuned, ouyang2022training, sanh2022multitask}, as depicted in Figure~\ref{figure:instruction_tuning}.
Instruction tuning is a supervised approach for teaching foundation models to follow instructions to solve new tasks, i.e., making output based on the instruction input. After instruction tuning, LLMs gain the ability to follow task instructions for new tasks without any demonstrated examples, thereby enhancing their generalization ability.

\textbf{\textit{In-context learning: zero-shot and few-shot.}}
In-context learning refers to the process of asking \llm{} to make inferences based on given demonstration examples or task descriptions~\cite{brown2020language}. It can be classified into two types: zero-shot learning and few-shot learning~\cite{brown2020language}. In zero-shot learning, LLMs generate responses solely based on natural language instruction descriptions, without any parameter updates. On the other hand, few-shot learning tasks involve LLMs making responses with only a few demonstration examples, again without updating model parameters.

\textbf{\textit{Parameter-efficient fine-tuning.}}
Fine-tuning the model on specific tasks effectively enhances its task-specific performance. However, the large number of parameters in the model makes it impractical to update all of them during fine-tuning. 
To address this issue, a Parameter-Efficient Fine-Tuning (PEFT) method~\cite{houlsby2019parameterefficient} is proposed, which maintains the best performance of the LLM while fine-tuning only a small number of parameters.
Several techniques fall under the umbrella of PEFT, including adapter tuning \cite{houlsby2019parameterefficient}, prefix tuning \cite{liu2022ptuning}, prompt tuning \cite{lester2021power}, and low-rank adaptation (\lora{}).
Among these techniques, \lora{} stands out for its ability to significantly reduce the number of parameters updated during fine-tuning by imposing a low-rank constraint on the updated matrix of each dense layer. 
Studies~\cite{hu2021lora} have shown superior performance of \lora{} compared to other PEFT methods while using fewer trainable parameters, making it a popular choice for efficient fine-tuning in open-source \llm{} implementations like \llama{}~\cite{touvron2023llama} and BLOOM~\cite{workshop2023bloom}.

\section{Experimental Setup}
This work makes the first attempt to comprehensively understand the performance of instruction-tuned LLMs on code comprehension and generation by answering the following research questions. 

\begin{itemize}[leftmargin=15pt]
    \item \textbf{RQ1 (Zero Shot):}  How do instructed LLMs perform on code comprehension and generation tasks in the zero-shot setting? 

    \item \textbf{RQ2 (Few Shot):} How do instructed LLMs perform on code comprehension and generation tasks in the few-shot setting? 

    \item \textbf{RQ3 (Fine Tuning):} How do instructed LLMs perform on code comprehension and generation tasks with further fine-tuning? 

    \item\textbf{RQ4 (Costs):} How are the costs of instructed  LLMs? 
\end{itemize}

RQ1 investigates the zero-shot generalization of instruction-tuned LLMs on different downstream code-relevant tasks;  RQ2 investigates the in-context learning capability of instruction-tuned LLMs in the few-shot setting as well as the impact of different shot-selection strategies; RQ3 investigates the performance of the instruction-tuned LLMs after being further fine-tuned on specific downstream task; and RQ4 compares the inference and fine-tuning costs of instructed LLMs in terms of the memory and time costs.

\subsection{Studied Tasks \& Datasets \& Metrics}
\subsubsection{Tasks}
In this work, we focus on four representative code comprehension and generation tasks that have been widely used in recent empirical studies on code intelligence~\cite{niu2023empirical, DBLP:conf/sigsoft/WangYGP0L22, zeng2022extensive}, namely Clone Detection (CD), Defect Detection (DD), Assertion Generation (AG), and Code Summarization (CS). 

\begin{itemize}[leftmargin=10pt]
    \item \textbf{Defect Detection (DD).} This task detects whether code snippets contain security-related vulnerabilities such as Denial of Service and injection. This task can reflect the model capability of understanding the code correctness.  
    
    \item \textbf{Clone Detection (CD).} This task identifies whether code snippets are in high similarity. This task can reflect the model capability of capturing the code similarity. 

    \item \textbf{Assertion Generation (AG).} This task generates the assertion statement for the given focal method (\eg{} the code under test) and the unit test prefix (\eg{} the test code prior to the assertion). This task can reflect the model capability of understanding the intention and generating oracles for the code under test. 
    
    \item \textbf{Code Summarization (CS).} This task generates a natural language description for the functionality of the given code snippet. This task can reflect the model capability of understanding the code intention and generating summaries in natural language.     
\end{itemize}

\subsubsection{Metrics}
For defect detection, clone detection, and assertion generation tasks, we follow the previous work~\cite{niu2022deep, niu2023empirical, lu2021codexglue} by using the same metrics. In particular, we adopt Accuracy (Acc) for defect detection, which calculates the ratio of the defects that are identified by the models; for clone detection, we adopt F1~\cite{f1}, which calculates the harmonic mean of precision and recall; for assertion generation, we adopt the exact match (EM), which calculates the ratio of generated assertions that are identical to the ground truth. 

\begin{figure}[htb]
    \centering
    \includegraphics[width=0.45\textwidth]{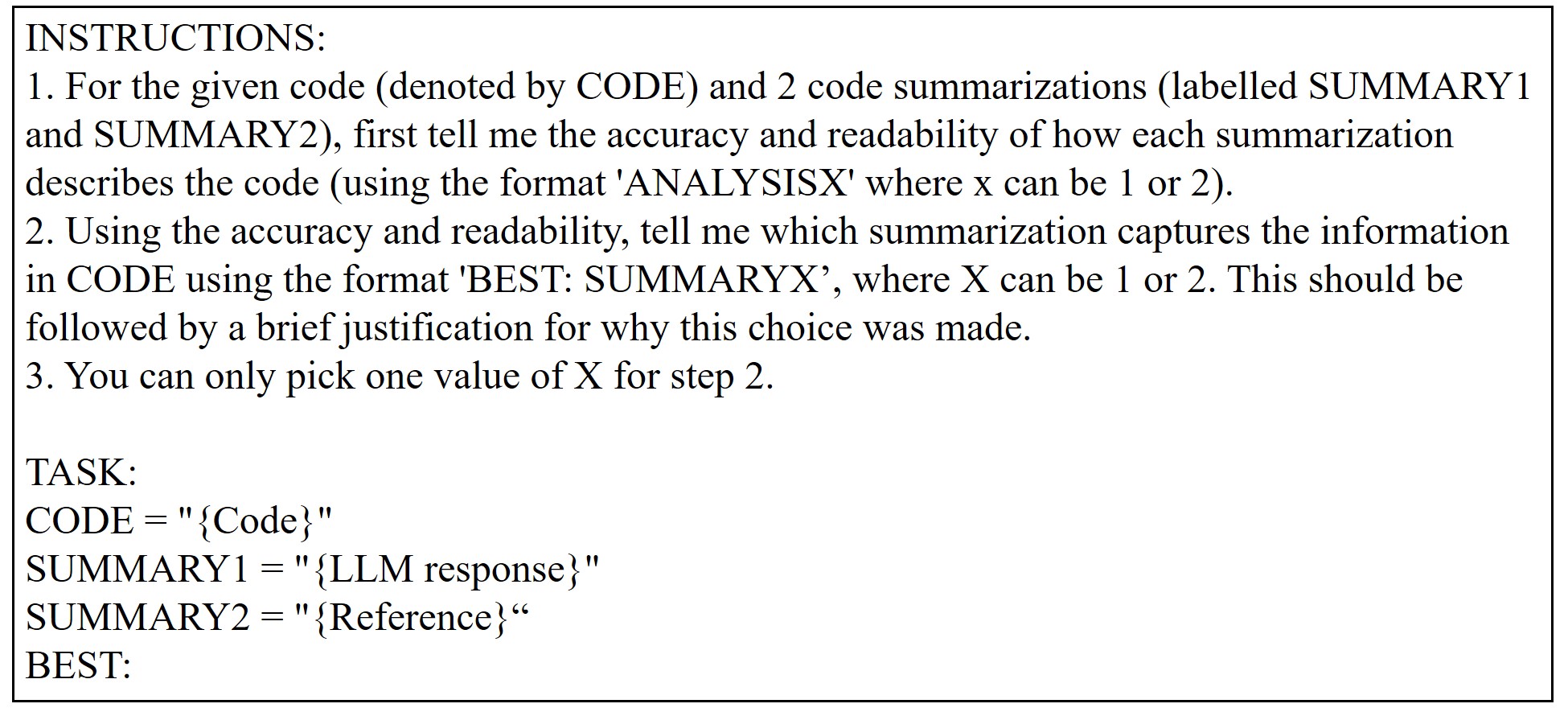}
    \caption{Prompt used for judgment}
    \label{figure:judgeprompt}
\end{figure}

For code summarization, we do not follow the previous work~\cite{niu2023empirical} to use the metric BLEU, which measures the similarity between the generated token sequence and the ground truth based on the n-gram matching. As pointed out by previous work on code summarization~\cite{DBLP:conf/sigsoft/RoyFA21}, BLEU fails to accurately capture the semantic similarity between token sequences, thus leading to inconsistent automated evaluation with manual assessment. In addition, such a bias would even be enlarged in evaluating instruction-tuned LLMs, since these models have been tuned on human-like instructions and tend to generate code summaries that exhibit different formats from the ground truth. Therefore, BLEU tends to  underestimate the similarity between the generated code summary and the ground truth. To overcome this issue, we follow the recent fashion in evaluating instruction-tuned LLMs by incorporating the powerful LLM ChatGPT~\cite{chatgpt} as the judge~\cite{zheng2023judging, gudibande2023false, wang2023large, bubeck2023sparks,chiang2023large,dettmers2023qlora,dubois2023alpacafarm,peng2023instruction}. 
In particular, we adopt the common prompt template used in previous work~\cite{zheng2023judging} to query ChatGPT to compare the quality of the generated code summary and the ground truth. Figure~\ref{figure:judgeprompt} presents an example for the prompt. In addition, to mitigate potential position bias, we further incorporate the few-shot strategy proposed in prior work~\cite{zheng2023judging}. In particular, we include two manually-crafted examples, labeled as ``SUMMARY1 good'' and ``SUMMARY2 good'' into the prompt to guide the evaluation process. We then calculate the \textit{preferred rate \preferrate} (\ie{} the frequency of cases that the generated summary is preferred by ChatGPT) to measure the model capability on the code summarization task. 

\subsubsection{Datasets}
Following previous work~\cite{niu2023empirical}, we adopt the widely-used datasets for each studied task. Considering the non-trivial costs of fine-tuning instructed LLMs and the extensive scale of our experiments, we do not use all the data items in the original datasets, but sample a portion of the data items as our training, validation, and testing datasets. Table~\ref{tab:evaluationtask} lists the detailed size of the datasets used in each task. In particular, we sample 2,000 items as the testing dataset for each task except the code summarization, given the large time and financial costs in querying ChatGPT API to calculate the metric \preferrate. 

\begin{table}[htb]
	\centering
	\small
	\caption{Summary of studied tasks, datasets and metrics}\label{tab:evaluationtask}
	
	\begin{adjustbox}{width=1.0\columnwidth}
	   	
	\begin{tabular}{l|c|c|c}
		\hline
        \textbf{Task} &   \textbf{Dataset} & \textbf{Train/Val/Test Size} & \textbf{Metric} \\ \hline
          
          \underline{\textbf{D}}efect \underline{\textbf{D}}etection (DD) & Devign~\cite{zhou2019devign} &  20k/2k/2k &	Acc \\
          \underline{\textbf{C}}lone \underline{\textbf{D}}etection (CD)	& BigCloneBench~\cite{6976121} & 100k/2k/2k & F1 \\ 

        \underline{\textbf{A}}ssert \underline{\textbf{G}}eneration (AG) & ATLAS~\cite{Watson_2020} & 120k/2k/2k & EM \\
	   \underline{\textbf{C}}ode \underline{\textbf{S}}ummarization (CS) & CodeSearchNet~\cite{husain2020codesearchnet} & 100k/2k/0.1k & \preferrate{} \\
        \hline

	\end{tabular}
	\end{adjustbox}
\end{table}

\begin{table*}[htb]
	\centering
	\small
	\caption{Summary of studied instruction-tuned \llm{}s } \label{table:models}

	\begin{adjustbox}{width=1.9\columnwidth}
	\begin{tabular}{l|c|c|c|c|c|c}
		\hline
  
        \multirow{2}{*}{\textbf{Instruction-tuned \llm{}}}
        & \multirow{2}{*}{\textbf{Foundation Model}} &
        \multicolumn{5}{c}{\textbf{Instruction-tuning Dataset}} \\ \cline{3-7}

        && \textbf{Domain} &\textbf{Dataset} &\textbf{Source} & \textbf{Size} &\textbf{Release Time}  \\ \hline
        
        {\chatglm{}-6B~\cite{zeng2022glm130b}} & GLM~\cite{chatglm}  & General & Chinese and English corpus~\cite{chatglm-corpus} & Unknown & Unknown & 2023/03\\
        {\alpaca{}-7B~\cite{alpaca}} & \llama{}~\cite{touvron2023llama} & General & Self-instruct~\cite{alpaca_data} & GPT-3 &  52k  &2023/03 \\
        {\vicuna{}-7B~\cite{vicuna2023}} & \llama{}~\cite{touvron2023llama}  & General & ShareGPT~\cite{sharegpt} & ChatGPT& 70k &2023/03 \\
        {\dolly{}-v2-7B~\cite{DatabricksBlog2023DollyV2}} & \pythia{}~\cite{biderman2023pythia}  & General & databricks-dolly-15k~\cite{dolly15k} & Manual& 15k &2023/04 \\
        {\stablelm{}-7B~\cite{stablelm}} & \stablelm{}-Base~\cite{stablelm-base} & General & Five conversational dataset~\cite{stablelm-corpus} &ChatGPT+GPT-3+Manual& Unknown&2023/04 \\
    
        {\codealpaca{}-7B}~\cite{codealpaca} & \llama{}~\cite{touvron2023llama} & Code & Code Alpaca~\cite{codealpaca_data}  & GPT-3 &  20K &2023/03 \\
        
        {\dolly{}-v2-12B}~\cite{DatabricksBlog2023DollyV2} & Pythia~\cite{biderman2023pythia} & General & databricks-dolly-15k~\cite{dolly15k} & Manual & 15k &2023/04 \\
        {\vicuna{}-13B}~\cite{vicuna2023} & \llama{}~\cite{touvron2023llama} & General & ShareGPT~\cite{sharegpt} & ChatGPT& 70k &2023/03 \\
        {\wizardcoder{}-15B}~\cite{luo2023wizardcoder} & StarCoder~\cite{li2023starcoder} & Code & CodeAlpaca with Evol-Instruct~\cite{evol-instruct} & GPT-3 & 78k & 2023/06 \\
        {\instructcodegen{}-16B}~\cite{instructcodegen} & \codegen{}-multi~\cite{nijkamp2022codegen} & Code & Code instructions~\cite{instructcodegencorpus}  &  Unknown & 250k &2023/05 \\
        



        \hline
	\end{tabular}
	\end{adjustbox}

\end{table*}

\vspace{-3mm}
\subsection{Studied Models}
\parabf{Studied Instruction-tuned LLMs.} Since instruction-tuned LLMs have been a recently emerging and rapidly developing domain (especially after the release of ChatGPT), our study mainly focuses on recent instruction-tuned LLMs that have been released in the past four months (\ie{} from March 1st to July 1st 2023). In particular, we exclude the instruction-tuned LLMs (i) that are the closed-source (\eg{} ChatGPT) due to lack of transparency and reproducibility, or (ii) that have more than 20B parameters (\eg{} Falcon 40B~\cite{falcon40b}) due to our resource constraints. As a result, our experiments include 10 recent instruction-tuned LLMs in total. 
Table~\ref{table:models} presents the details of our studied instruction-tuned LLMs. From the table, our experiments cover a wide spectrum of instruction-tuned LLMs that are diverse in multiple dimensions, such as model sizes, foundation models, and instruction data domains. For example, the sizes of our studied instruction-tuned LLMs range from 6B (\eg{} \chatglm{}-6B) to 16B (\eg{} \instructcodegen{}-16B), and our studied instruction-tuned LLMs are built on different foundation models (\eg{} \llama{}, \pythia{}, and GLM). Furthermore, we include both LLMs that are (i) tuned on general instructions or (ii) tuned on code-specific instructions (\ie{} instructions are code-related tasks). For example, \alpaca{} is built by tuning the foundation model \llama{} on a general instruction dataset~\cite{alpaca_data}, while \codealpaca{} is built by tuning the foundation model \llama{} on the code instruction dataset~\cite{codealpaca_data}. 

\parabf{Other Baselines.} In addition to the 10 instruction-tuned LLMs mentioned above, we further include the five baselines as follows. 
(i) \textit{Small SOTA models:} we include four small pre-trained models (with far less than 1B parameters) that have achieved the best performance on our studied tasks after being fine-tuned on each task as shown in the latest work~\cite{niu2023empirical}, that is, CodeGPT-adapted~\cite{codegpt} for clone detection, CoText~\cite{phan2021cotext} for defect detection, PLBART~\cite{ahmad-etal-2021-unified} for assertion generation, and CodeT5~\cite{wang2021codet5} for code summarization.
It is worth noting that we use second-ranked models for the defect detection and clone detection tasks since the Top-1 model SynCoBERT~\cite{wang2021syncobert} is closed-source.
We include these models so as to enable the comparison between instruction-tuned LLMs and these small SOTA pre-trained models.
(ii) \textit{Large models without instruction tuning:} we also include one large language model that is in a similar scale range as our studied instruction-tuned LLMs but without instruction tuning, \ie{} \codegen{}-6B~\cite{nijkamp2022codegen}, which is pre-trained on a dataset of multiple programming languages in an autoregressive manner. We include this model so as to compare similar-scaled LLMs with/without instruction tuning.

\subsection{Prompt Design}\label{sec:setup:prompt}

\begin{table}
	\centering
	\small
	\caption{Prompt constitution}\label{table:prompt}
	
	\begin{adjustbox}{width=1.0\columnwidth}
	   	
	\begin{tabular}{c|l}
		\hline
        \textbf{LLM} & \textbf{Prompt} \\ \hline  
        
        {\vicuna{}} & {\$\{\textit{System prompt}\}. \textbf{User}: \$\{\textit{Task instruction}\}. \textbf{Assistant}:}  \\ \hline

        {\stablelm} & {\$\{\textit{System prompt}\}. \textbf{$<|$USER$|>$}:\$\{\textit{Task instruction}\}.  \textbf{$<|$ASSISTANT$|>$}:} \\ \hline
        
        {\alpaca{}} & \multirow{4}{*}{\${\{\textit{System prompt}\}. \textbf{Instruction}: \$\{\textit{Task instruction}\}. \textbf{Response}:}} \\

        {\dolly{}} & ~  \\ 
                
        {Instruct-\codegen{}} & ~  \\ 
        
        {\wizardcoder{}} & ~  \\ \hline

        {\chatglm{}} & \multirow{2}{*}{{\textbf{Instruction}: \$\{\textit{Task instruction}\}. \textbf{Response}:}} \\ 
    
        {\codealpaca{}} & ~  \\ \hline

	\end{tabular}
	\end{adjustbox}
\end{table}

Typically, the prompt for querying instruction-tuned LLMs comprises three parts: a system prompt, a task instruction, and the keywords to separate different parts. Table~\ref{table:prompt} presents the prompt constitution of each model. We then explain each part in detail.

\textit{The system prompt} is the starting sentence in the prompt, which is often defined by each instruction-tuned LLMs itself. For example, \alpaca{} officially suggests using the sentence ``\textit{Below is an instruction that describes a task, paired with an input that provides further context. Write a response that appropriately completes the request.}'' as its system prompt. Note that the system prompt is optional since not all instruction-tuned LLMs have incorporated the system prompt (\eg{} \chatglm{} and \codealpaca{}). Thus, our experiments only include the system prompt for the instruction-tuned LLMs whose official documentation or code repositories have provided its system prompt. More details on the system prompt used in each model can be found in our website~\cite{experimentdata}.

\textit{The task instruction} describes the detailed task that needs to be done by \llm{}s. Following the prompt design principle that the task goal should be clarified~\cite{zhao2023survey} and referring to the task descriptions in prior studies~\cite{lu2021codexglue, niu2022deep}, we design the task instruction for each task as follows. For defect detection, the task instruction is ``\textit{Is there a defect in \$\{code\}, and respond to YES or NO}'', where \textit{\$\{code\}} is the given code snippet for defect detection. For clone detection, the task instruction is ``\textit{{Is there a clone relation between the \$\{code1\} and \$\{code2\}, and respond to YES or NO}}'', where \textit{\$\{code1\}} and \textit{\$\{code2\}} are the given pair of code snippets for clone detection. For assertion generation, the task instruction is ``\textit{Generate an assertion code at the $<$AssertPlaceHolder$>$ in the following \textit{\$\{code\}} using Junit API}'', where \textit{\$\{code\}} is the given focal method (the method under test) and the test prefix and $<$AssertPlaceHolder$>$ is a placeholder in \textit{\$\{code\}}. For code summarization, the task instruction is ``\textit{Generate the method-level comment for the following {\$\{code\}}}'', where \textit{\$\{code\}} is the given code snippet for summarization. 

\textit{The Keywords} are the reserved tokens defined by instruction-tuned LLMs themselves to indicate different parts in the prompt. For example, \vicuna{} uses the keyword ``\textit{User}'' to indicate the following task instruction while \stablelm{} uses ``\textit{$<|$USER$|>$}''. The detailed keywords defined by each model are highlighted in bold in Table~\ref{table:prompt}.

\subsection{Experimental Procedure}~\label{sec:setup:procedure}
In RQ1, we investigate the zero-shot generalization of instruction-tuned LLMs on each code comprehension and generation task. 
In particular, for all the studied 10 instruction-tuned LLMs as well as the uninstructed baseline \codegen{}-6B, we directly query them via the prompt (defined in Section~\ref{sec:setup:prompt}). For the four small SOTA models, we first fine-tune them on each downstream task and compare their fine-tuned performance on the testing dataset, since such small and uninstructed models are not applicable in the zero-shot setting.

In RQ2, we investigate the few-shot performance as well as the impact of different shot selection strategies.
We currently focus on one shot, since two or more shots sometimes exceed the maximum tokens taken by the model on our tasks. In particular, we include one demonstration example at the beginning of the prompt in the one-shot setting. We consider three common shot-selection strategies that have been widely used in previous work~\cite{nashid2023retrieval}, including (i) \underline{F}ix \underline{O}ne (\textit{FO}) strategy which consistently uses a same manually-designed example for all the testing data items in each task;  (ii)  \underline{R}andomly-selected \underline{O}ne (\textit{RO}) which randomly selects a demonstration example from the training dataset for each testing data item in each task; and (iii) \underline{B}M25-based \underline{O}ne (\textit{BO}), which retrieves the demonstration example with the highest BM25~\cite{bm25} similarity as the testing data item from the training dataset.

In RQ3, we fine-tune all the studied models on the same training dataset for each downstream task. In particular, for large models such as 10 instruction-tuned LLMs and \codegen{}-6B, we fine-tune them with the parameter-efficient tuning strategy \textit{Low-Rank Adaptation} (\lora{}), which accelerates the training of large models with fewer memory costs by including a small number of trainable parameters (\ie{} adapters) and freezing the original parameters for LLM fine-tuning. 
For small SOTA models, we follow the previous work~\cite{niu2023empirical} to tune all of their parameters. 

In RQ4, we record the average time costs and the memory costs of each model during its training and inference phases. 

In total, this work performs large-scale experiments for 15 models on 4 code-relevant tasks in diverse settings (\ie{} zero-shot, few-shot with three different shot-selection strategies, and fine-tuning), which takes around 1000 GPU-hours on one NVIDIA A800-80GB GPU.

\subsection{Implementation Details}

\parabf{Models implementation.} For all the studied large models, we directly download them from their official code repositories, since they are open-source. In addition, we directly use the fine-tuning and inference scripts of these models if they are provided in the repositories. For the compared small SOTA models, since the results reported in previous work~\cite{niu2023empirical} are based on the training dataset and testing dataset that are different from our experiments, for a fair comparison, we fine-tune these models by ourselves on the same training dataset used to fine-tune the instruction-tuned models. In particular, we strictly follow the same hyper-parameters of training these SOTA models as previous work~\cite{niu2023empirical}, including the batch size, training epochs, and learning rate. 

\parabf{Hyper-parameters.} For all studied LLMs, we set ``max\_length'' to 2,048 tokens. To minimize the randomness in the generated responses, we set parameter ``do\_sample'' to False to ensure deterministic results. The other hyper-parameters of instruction-tuned LLMs are set as the default values provided in their code repositories.

\parabf{Environment.} 
Our experiment is conducted on a single NVIDIA A800-80G GPU. The operating system is Ubuntu 20.04.6 LTS.

\vspace{-3mm}
\section{Experiment Results}
\subsection{Performance in Zero Shot (RQ1)}

Table~\ref{table:multiTask} presents the performance of studied instruction-tuned LLMs on each code-relevant task in the zero-shot setting. We also present the results of relatively-smaller pre-trained models that have been fine-tuned on each task (the ``SOTA Model'' Row) and the results of the code large language model without instruction tuning (the ``\codegen{}-6B'' Row) for comparison. Note that the performance of SOTA models reported in our work is lower than that in the previous work~\cite{niu2023empirical} since they are trained and evaluated on a smaller training dataset in our work. The best performance in each task is highlighted. Based on the table, we have the following observations.

\begin{table}[htb]
	\centering
	\caption{Performance in zero-shot}~\label{table:multiTask}
 
	\begin{adjustbox}{width=0.87 \columnwidth}
 \begin{threeparttable}
	\begin{tabular}{c|c|c|c|c}
		\hline
        \textbf{Models} & \textbf{DD (\%)}  & \textbf{CD (\%)} &  \textbf{AG (\%)} &  \textbf{CS (\%)}  \\ \hline
        SOTA Model & \textbf{\underline{58.7}} & 7.4 & \textbf{\underline{25.7}} & 24.0 \\ \hline  
        \codegen{}-6B  & 0.3 & 1.4 & 0.0 & 0.0  \\ \hline \hline 
        \chatglm{}-6B & 7.1 & 17.5 & 1.7 & 45.0   \\ \hline
        \vicuna{}-7B & 54.0 & 13.2 & 10.1 & 48.0  \\ \hline
        \alpaca{}-7B & 45.8 & 22.1 & 5.3 & 32.0 \\ \hline
        \dolly{}-7B &  33.1 & 21.3 & 1.9 & 12.0 \\ \hline
        \stablelm{}-7B & 44.3 & \textbf{\underline{24.3}} & 1.1 & 30.0  \\ \hline
        \codealpaca{}-7B & 51.9 & 1.4 & 4.4 & 9.0 \\ \hline
        \dolly{}-12B & 33.8 & 23.5 & 1.0 & 5.0   \\ \hline
        \vicuna{}-13B & 49.8 & 14.1 & 12.0 & 63.0 \\ \hline
        \wizardcoder{}-15B & 54.4 & 23.8 & 19.4 & \textbf{\underline{71.0}} \\ \hline
        Instruct-\codegen{}-16B & 47.8 & 14.2 & 8.4 & 9.0 \\ \hline
        
	\end{tabular}
    \end{threeparttable}
	\end{adjustbox}
\end{table}

First, even in the zero-shot setting, instruction-tuned LLMs surprisingly achieve comparable sometimes even better performance on code comprehension and generation tasks compared to SOTA small models which have been specifically fine-tuned on each downstream task.
For example, in clone detection and code summarization, most instruction-tuned LLMs outperform the SOTA fine-tuned models. 
Additionally, in defect detection and assertion generation, although the SOTA fine-tuned models hold the Top-1 performance on these two tasks, some instruction-tuned LLMs (\eg{} \wizardcoder{}-15B) are able to achieve comparable performance (\ie{} 54.4\% v.s. 58.7\% on defect detection). 
It is notable that \codegen{}-6B (an LLM with a comparable number of parameters but without instruction tuning) performs extremely poorly in such a zero-shot setting since it always tends to generate code snippets unrelated to the task. It confirms that instruction tuning indeed improves the zero-shot generalization of LLMs on unseen code-related tasks.

Second, interestingly, we find that LLMs tuned by code-specific instructions do not necessarily outperform LLMs tuned by general instructions. 
For example, \alpaca{}-7B outperforms \codealpaca{}-7B on all the code-relevant tasks except the defect detection, while they are basically tuning the same foundation model \llama{} on the general instructions or code-specific instructions respectively. 
In addition, we observe that the smaller general instruction-tuned LLMs (\eg{} \vicuna{}-7B) have the chance to outperform the larger code-specific instruction-tuned LLMs (\eg{} \instructcodegen{}-16B) on some tasks. 
One potential reason might be that the general instructions (\eg{} ShareGPT~\cite{sharegpt}) also contain code-specific instructions and the other code-unrelated instructions can further be helpful for improving the instruction understanding capability of LLMs.
In summary, general instruction-tuned \llm{}s can achieve competitive performance on code comprehension and generation tasks. 

Third, we observe that increasing the model scale sometimes brings marginal improvements or sometimes even decreases in the zero-shot setting. For example, \dolly{} performs much worse on assertion generation and code summarization when the model scale increases from 7B to 12B. In addition, for \vicuna{}, although enlarging the model scale from 7B to 13B improves the performance on generation tasks such as assertion generation and code summarization, the impact on classification tasks like defect detection or clone detection is rather small. In fact, previous studies~\cite{peinl2023evaluation, kim2023aligning} also reveal similar observations that Dolly-12B performs worse than Dolly-6B in generative tasks, and ALMoST-7B~\cite{kim2023aligning} outperforms Alpaca-13B, Dolly-12B, and OpenAssistant-12B~\cite{köpf2023openassistant}.

\finding{In the zero-shot setting, instruction-tuned LLMs are competitive and sometimes even better on code comprehension and generation tasks compared to  small SOTA models specifically fine-tuned on each downstream task. In addition, code-instruction-tuned LLMs do not necessarily outperform general-instruction-tuned LLMs on code comprehension and generation tasks, and the improvement from the larger model scale sometimes can be limited or even negative.}

\begin{table}[htb]
	\centering
	\caption{\llm{} performance under zero-shot and one-shot}\label{table:zerovsone}
 
	\begin{adjustbox}{width=0.97 \columnwidth}
 \begin{threeparttable}
	\begin{tabular}{c|c|c|c|c|c|c|c|c}
		\hline
        \multirow{2}{*}{\textbf{\llm{}}} & \multicolumn{2}{c}{\textbf{DD (\%)}}  &  \multicolumn{2}{|c}{\textbf{CD (\%)}} &  \multicolumn{2}{|c}{\textbf{AG (\%)}} &  \multicolumn{2}{|c}{\textbf{CS (\%)}}  \\ \cline{2-9}  
        ~ & ZO & BO & ZO & BO & ZO & BO & ZO & BO  \\ \hline
        \codegen{}-6B  & 0.3 & 43.6 & 1.4 & 23.4 & 0.0 & 56.2 & 0.0 & 13.0   \\ \hline
        \chatglm{}-6B & 7.1 & 54.2 & 17.5 & 12.8 & 1.7 & 46.2 & 45.0 & \textbf{\underline{54.0}}  \\ \hline
        \vicuna{}-7B & 54.0 & 54.1 & 13.2 & - & 10.1 & 31.2 & 48.0 & 37.0  \\ \hline
        \alpaca{}-7B & 45.8 & \textbf{\underline{55.4}} & 22.1 & - & 5.3 & 41.4 & 32.0 & 6.0  \\ \hline
        \dolly{}-7B & 33.1 & 49.9 & 21.3 & \textbf{\underline{23.5}} & 1.9 & 51.0 & 12.0 & 14.0 \\ \hline
        \stablelm{}-7B & 44.3 & 43.4 & \textbf{\underline{24.3}} & - & 1.1 & 44.4 & 30.0 & 19.0  \\ \hline
        \codealpaca{}-7B & 51.9 & 50.3 & 1.4 & 10.3 & 4.4 & 35.1 & 9.0 & 34.0 \\ \hline
        \dolly{}-12B & 33.8 & 52.7 & 23.5 & 22.6 & 1.0 & 51.7 & 5.0 & 8.0  \\ \hline
        \vicuna{}-13B & 49.8 & 53.0 & 14.1 & 6.5 & 12.0 & 44.0 & 63.0 & 24.0 \\ \hline
        \wizardcoder{}-15B & \textbf{\underline{54.4}} & 53.8 & 23.8 & 7.3 & \textbf{\underline{19.4}} & \textbf{\underline{63.3}} & \textbf{\underline{71.0}} & 50.0 \\ \hline
        Instruct-\codegen{}-16B & 47.8 & 54.6 & 14.2 & 20.7 & 8.4 & 55.0 & 9.0 & 41.0 \\ \hline
        
	\end{tabular}
    \end{threeparttable}
	\end{adjustbox}
\end{table}

\begin{table*}[htb]
	\centering
	\caption{Performance with different one-shot selection strategies }~\label{table:multione}
	
	\begin{adjustbox}{width=1.47 \columnwidth}
 \begin{threeparttable}
	\begin{tabular}{c|c|c|c|c|c|c|c|c|c|c|c|c}
		\hline
        \multirow{2}{*}{\textbf{\llm{}}} & \multicolumn{3}{c}{\textbf{DD (\%)}}  &  \multicolumn{3}{|c}{\textbf{CD (\%)}} &  \multicolumn{3}{|c}{\textbf{AG (\%)}} &  \multicolumn{3}{|c}{\textbf{CS (\%)}}  \\ \cline{2-13}  
        ~ & FO & RO & BO & FO & RO & BO & FO & RO & BO & FO & RO & BO  \\ \hline
        \codegen{}-6B  & 42.3 & 40.6 & \textbf{\underline{43.6}} & \textbf{\underline{24.3}} & 23.8 & 23.4 & 5.5 & 7.4 & \textbf{\underline{56.2}} & 10.0 & 6.0 & \textbf{\underline{13.0}} \\ \hline
        \chatglm{}-6B & 45.6 & 50.9 & \textbf{\underline{54.2}} & 6.7 & 11.3 & \textbf{\underline{12.8}} & 2.6 & 3.9 & \textbf{\underline{46.2}} & 44.0 & 35.0 & \textbf{\underline{54.0}} \\ \hline
        \vicuna{}-7B & 53.4 & 53.9 & \textbf{\underline{54.1}} & - & - & - & 8.4 & 9.5 & \textbf{\underline{31.2}} & \textbf{\underline{65.0}} & 26.0 & 37.0 \\ \hline
        \alpaca{}-7B & 46.3 & 50.1 & \textbf{\underline{55.4}} & - & - & - & 0.0 & 1.0 & \textbf{\underline{41.4}} & 2.0 & 2.0 & \textbf{\underline{6.0}} \\ \hline
        \dolly{}-7B & 49.7 & 46.0 & \textbf{\underline{49.9}} & 16.2 & - & \textbf{\underline{23.5}} & 4.7 & 5.2 & \textbf{\underline{51.0}} & \textbf{\underline{15.0}} & 13.0 & 14.0 \\ \hline
        \stablelm{}-7B & 44.9 & \textbf{\underline{47.4}} & 43.4 & - & \textbf{\underline{0.7}} & - & 0.9 & 1.2 & \textbf{\underline{44.4}} & 16.0 & 8.0 & \textbf{\underline{19.0}} \\ \hline
        \codealpaca{}-7B & 50.5 & \textbf{\underline{51.5}} & 50.3 & 6.4 & \textbf{\underline{18.2}} & 10.3 & 4.6 & 3.1 & \textbf{\underline{35.1}} & 27.0 & 26.0 & \textbf{\underline{34.0}} \\ \hline
        \dolly{}-12B & 41.3 & 44.3 & \textbf{\underline{52.7}} & 23.6 & \textbf{\underline{24.3}} & 22.6 & 8.8 & 7.0 & \textbf{\underline{51.7}} & 6.0 & 5.0 & \textbf{\underline{8.0}} \\ \hline
        \vicuna{}-13B & \textbf{\underline{53.9}} & 52.4  & 53.0 & 1.4 & \textbf{\underline{9.7}} & 6.5 & 14.1 & 11.6 & \textbf{\underline{44.0}} & \textbf{\underline{42.0}} & 23.0 & 24.0 \\ \hline
        \wizardcoder{}-15B & 53.6 & \textbf{\underline{53.8}} & \textbf{\underline{53.8}} & - & - & \textbf{\underline{7.3}} & 25.5 & 23.1 & \textbf{\underline{63.3}} & \textbf{\underline{72.0}} & 53.0 & 50.0 \\ \hline
        Instruct-\codegen{}-16B & 50.9 & 50.4 & \textbf{\underline{54.6}} & \textbf{\underline{24.3}} & 23.2 & 20.7 & 9.2 &	9.0 & \textbf{\underline{55.0}} & 28.0 & 29.0 & \textbf{\underline{41.0}} \\ \hline
        
	\end{tabular}
    \end{threeparttable}
	\end{adjustbox}
\end{table*}

\subsection{Performance in Few Shot (RQ2)}
In RQ2, we extensively investigate instruction-tuned LLMs in the few-shot setting (one-shot in our experiments), including the performance and instruction-following capability comparison to the zero-shot setting (in Section~\ref{sec:rq2:comparison} and Section~\ref{sec:rq2:instructfollow}) and the impact of different shot selection strategies (Section~\ref{sec:rq2:strategy}).

\vspace{-2mm}
\subsubsection{Comparison to zero-shot performance}~\label{sec:rq2:comparison}
Table~\ref{table:zerovsone} presents the performance of instruction-tuned LLMs in the zero-shot setting (\ie{} in ``ZO'' column) and in the one-shot setting  (\ie{} in ``BO'' column). For space limits, here we only present the results of the best one-shot selection strategy (\ie{} the BM25-based one) and the comparison among different strategies can be found in Section~\ref{sec:rq2:strategy}. Based on the table, we have the following observations.

Overall, including one demonstration example in the input improves the performance of all the studied instruction-tuned LLMs on most tasks, \ie{} 25 out of 40 cases (10 instruction-tuned LLMs $\times{}$ 4 tasks). For example, on assertion generation, including an example substantially improves the performance of all the instruction-tuned LLMs. In particular, for those cases that the instruction-tuned \llm{}s perform extremely poor (\ie{} with less than 10\% performance), adding one shot could always improve the performance by a large margin. For example, the performance of \chatglm{}-6B increases from 7.1\% to 54.2\% on defect detection, and the performance of \instructcodegen{}-16B increases from 9.0\% to 41.0\% on code summarization. The improvements mentioned above indicate the strong in-context learning capability of the instruction-tuned LLMs. Such an observation is consistent as previous evaluation of instruction-tuned LLMs on other tasks such as coreference resolution, closed-book QA and relation extraction  ~\cite{ye2023incontext,wei2022finetuned,labrak2023zeroshot}.

\finding{Including demonstration examples substantially improves the performance of instruction-tuned LLMs on most code comprehension and generation tasks; and for the cases that instruction-tuned LLMs perform extremely poor in the zero-shot setting, including one shot consistently improves their performance in an especially-large margin.}

\begin{figure}[htb]
    \centering
    \includegraphics[width=0.5\textwidth]{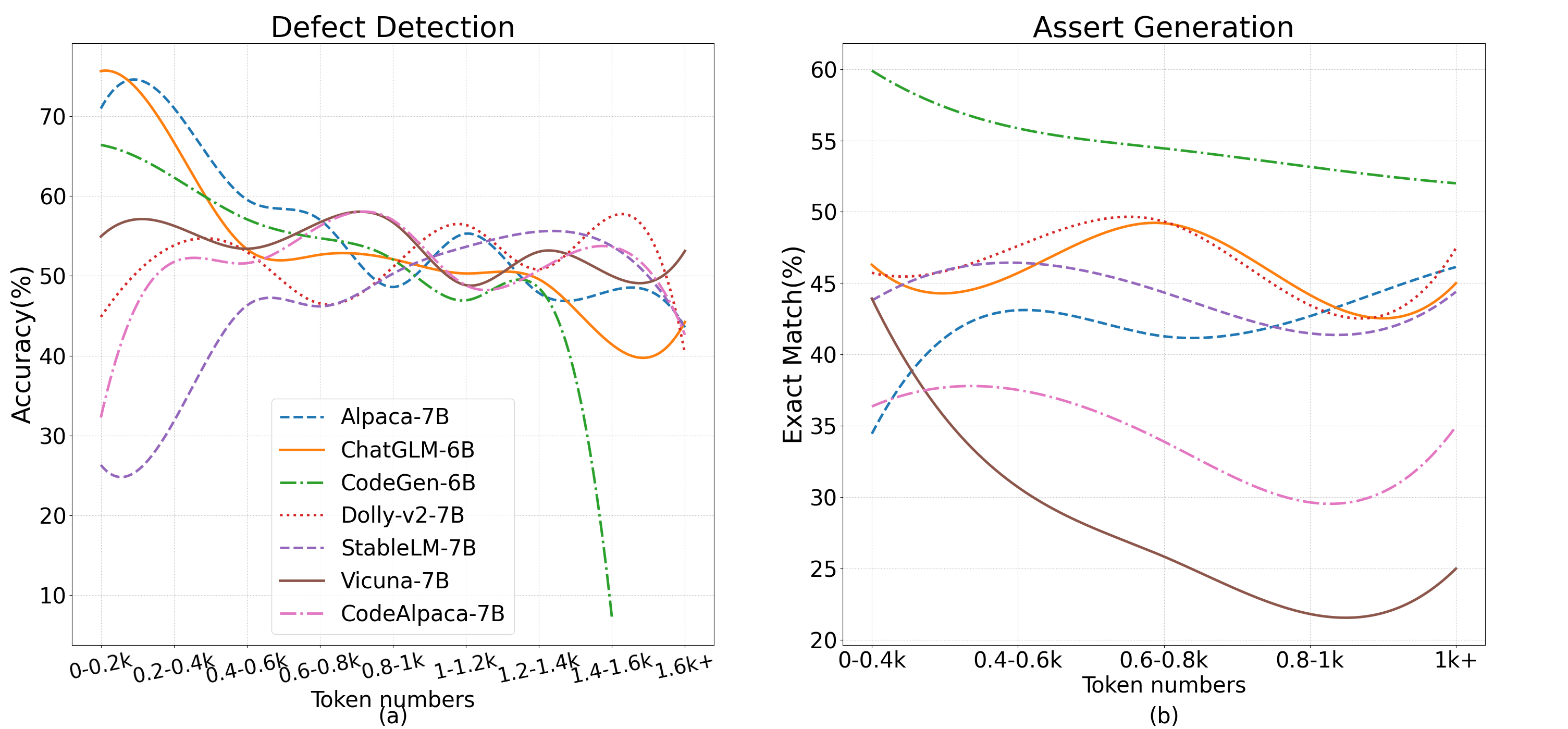}
    \caption{Model performance with different input lengths}
    \label{figure:promptlength}
\end{figure}

On the other hand, there are a few cases that adding examples causes unstable and even worse performance of instruction-tuned LLMs. For example, on clone detection, multiple instruction-tuned LLMs (\ie{} \vicuna{}-7B, \alpaca{}-7B, and \stablelm{}-7B) exhibit biased responses, \ie{} all their binary classification predictions are the same (all yes or all no) on the entire testing dataset, leading to invalid calculation of the metric F1 (marked as ``-'' in the table). Although we have adjusted the prompt with many attempts, the same observation persists. In fact, we are not the first work to find some unintuitive model behaviors in the few-shot setting. For example, previous work~\cite{DBLP:conf/emnlp/MinLHALHZ22, DBLP:conf/icml/ZhaoWFK021} shows that the performance of LLM is impacted by the order of examples in the input while sometimes even incorrect examples are helpful for LLM in the few-shot setting. In addition to unstable inherency of the few-shot learning, recent work~\cite{DBLP:journals/corr/abs-2307-03172} shows that in the general domain LLMs sometimes perform much worse when the input context becomes longer. Compared to the zero-shot setting, the input length is much longer in the one-shot setting given the inclusion of additional examples, which thus sometimes distracts the model and degrades the performance. To validate this assumption, we further present the model performance with inputs of different lengths in Figure~\ref{figure:promptlength}. For space limits, we mainly present the results on defect detection and assertion generation tasks. In particular, the x-axis presents the number of tokens (each LLM uses its own associated tokenizer) while the y-axis presents the corresponding performance. Overall, for most models, we observe a decreasing trend in the model performance with the increasing length of the inputs, implying that models struggle to utilize useful information in the long inputs. Our results extend the previous finding on general domain~\cite{DBLP:journals/corr/abs-2307-03172} that the performance of instructed LLMs also decreases with the longer inputs on code-specific tasks. 

\finding{In the one-shot setting, adding examples sometimes causes unstable and even worse performance of instruction-tuned LLMs. One potential reason might be the limited capability of existing instruction-tuned LLMs on utilizing longer input contexts.  }

\begin{table}[htb]
	\centering
	\caption{Instruction-following capability}\label{table:instructioncompare}
 
	\begin{adjustbox}{width=0.87 \columnwidth}
 \begin{threeparttable}
	\begin{tabular}{c|c|c|c|c|c|c}
		\hline
        \multirow{2}{*}{\textbf{\llm{}}} & \multicolumn{2}{c}{\textbf{DD (\%)}}  &  \multicolumn{2}{|c}{\textbf{CD (\%)}} &  \multicolumn{2}{|c}{\textbf{AG (\%)}}   \\ \cline{2-7}  
        ~ & ZO & BO & ZO & BO & ZO & BO   \\ \hline
        \codegen{}-6B  & 0.4 & 77.8 &  0.3 & 94.2 & 0.0 & 96.6  \\ \hline
        \chatglm{}-6B & 16.2 & 98.2 & 32.4 & 95.3 & 98.7 & 99.7   \\ \hline
        \vicuna{}-7B & 100.0 & 100.0 & 100.0 & 100.0 & 82.8 & 99.8   \\ \hline
        \alpaca{}-7B & 99.3 & 100.0 & 100.0 & 100.0 & 97.5 & 100.0   \\ \hline
        \dolly{}-7B & 65.6 & 95.7 & 61.0 & 97.8 & 13.6 & 91.3  \\ \hline
        \stablelm{}-7B & 92.5 & 96.6 & 90.6 & 98.2 & 27.9 & 99.6   \\ \hline
        \codealpaca{}-7B & 97.4 & 96.3 & 100.0 & 100.0 & 6.4 & 94.0  \\ \hline
        \dolly{}-12B & 65.2 & 93.8 & 98.9 & 100.0 & 25.0 & 98.6   \\ \hline
        \vicuna{}-13B & 92.0 & 99.2 & 61.2 & 100.0 & 99.4 & 99.2   \\ \hline
        \wizardcoder{}-15B & 98.7 & 98.2 & 100.0 & 100.0 & 95.4 & 99.5  \\ \hline
        Instruct-\codegen{}-16B & 97.7 & 97.6 & 99.7 & 97.7 & 56.0 & 99.8  \\ \hline
        
	\end{tabular}
    \end{threeparttable}
	\end{adjustbox}
\end{table}

\subsubsection{Instruction-following Capability}~\label{sec:rq2:instructfollow}
We further compare the model capability of following instructions in the zero-shot and one-shot settings. The model is considered as following the given instruction, when its response adheres to the requirement in instructions. For example, in classification tasks (\eg{} clone detection and defect detection), an instruction-following response is supposed to answer either ``YES'' or ``NO''; for the generation tasks (\eg{} assertion generation), an instruction-following response is supposed to return an assertion statement. The main focus is  on the format rather than the correctness of the model response. Instruction following capability is an essential indicator in task generalization and has been extensively evaluated in previous work~\cite{DBLP:journals/corr/abs-2307-10558}. 

Table~\ref{table:instructioncompare} shows the instruction-following capability of each model in both zero-shot and one-shot settings. First, in the zero-shot setting, we find that most instruction-tuned LLMs exhibit a high instruction-following capability on the code comprehension and generation tasks; and the model without instruction tuning (\eg{} \codegen{}-6B) exhibits extremely poor instruction-following capability. The results confirm that instruction tuning indeed improves instruction-following capability of models. Second, in the one-shot setting, we find that adding a demonstration example to the input substantially improves the instruction-following capability. For example, on defect detection, the instruction-following capability of \chatglm{}-6B improves from 16.2\% to 98.2\%. The increasing instruction-following capability from the zero-shot setting to the one-shot setting might also be the reason for the performance improvement in the one-shot setting observed in Section~\ref{sec:rq2:comparison}. 

\finding{Instructed LLMs exhibit a reasonable instruction-following capability on code comprehension and generation tasks in the zero-shot setting, and incorporating a demonstration example further effectively improves their instruction-following ability, which potentially helps improve the performance on downstream tasks.}

 
        

\begin{table}[htb]
	\centering
	\caption{P-values in paired T-test at significance level of {0.05}}~\label{table:pvalue}
 
	\begin{adjustbox}{width=0.8 \columnwidth}
 \begin{threeparttable}
	\begin{tabular}{c|c|c|c|c}
	\hline
        \textbf{Control Group} & \textbf{DD}  & \textbf{CD} &  \textbf{AG} &  \textbf{CS}  \\ \hline
        FO/BO & 0.0571 & 0.0857 &  \cellcolor[RGB]{225,225,225}{2.5089e-08} & 0.5661 \\ \hline  
        RO/BO  & 0.0643 & 0.5702 &  \cellcolor[RGB]{225,225,225}{1.4141e-08} &  \cellcolor[RGB]{225,225,225}{0.0053}  \\ \hline 
        FO/RO  & 0.3374 & 0.7275 & 0.6697 &  \cellcolor[RGB]{225,225,225}{0.0314}  \\ \hline 
        
	\end{tabular}
    \end{threeparttable}
	\end{adjustbox}
\end{table}

\subsubsection{\llm{} Performance with different one-shot selection strategies}~\label{sec:rq2:strategy}
Table~\ref{table:multione} shows the performance of instruction-tuned LLMs with different one-shot selection strategies. As mentioned in Section~\ref{sec:setup:procedure}, we consider three selection strategies, including the fix one shot (``FO'' column), the randomly-selected one shot (``RO'' column), and the BM25-similarity-based one shot (``BO'' column). Similarly, the ``-'' indicates that the LLM makes biased prediction on the entire testing set and thus leads to an invalid calculation of F1 metric on the clone detection task. In addition, we further conduct a Paired T-Test~\cite{hsu2014paired} to check whether there is a statistically significant difference between each strategy at the significance level of 0.05. Table~\ref{table:pvalue} presents the p-values between each two strategies, and the cases with significant difference (p-value $<$ 0.05) are highlighted. 

Based on Table~\ref{table:multione} and Table~\ref{table:pvalue}, we have the following two observation. First, on generation tasks (\ie{} assertion generation and code summarization), the BM25-based strategy is the best shot-selection strategy significantly better than random selection and fixed selection. Such an observation is consistent as previous work on the few-shot selection strategy for the close-source instruction-tuned LLM Codex~\cite{codex}, which also finds the BM25-based selection is better than random selection for Codex on assertion generation and program repair tasks. Our results further confirm that the superior of BM25-based selection strategy for a wide spectrum of open-source instruction-tuned LLMs on assertion generation and code summarization. Second, interestingly, on classification problems (\ie{} defect detection and clone detection), we find that the widely-used BM25-based has no significant difference from the other two basic strategies.

\finding{In the one-shot setting,  BM25-based selection strategy is often the best on generation problems such as assertion generation and code summarization, but has no  significant difference than basic strategies (random or fixed) on classification problems such as defect detection and clone detection.}


\begin{table*}[htb]
	\centering
	\caption{Fine-tuned \llm{} performance comparison with zero-shot and one-shot (+/- value against the FT)}~\label{table:tunedResult} 
	
	\begin{adjustbox}{width=1.57 \columnwidth}
 \begin{threeparttable}
	\begin{tabular}{c|c|c|c|c|c|c|c|c|c|c|c|c}
		\hline
        \multirow{2}{*}{\textbf{\llm{}}} & \multicolumn{3}{c}{\textbf{DD (\%)}}  &  \multicolumn{3}{|c}{\textbf{CD (\%)}} &  \multicolumn{3}{|c}{\textbf{AG (\%)}} &  \multicolumn{3}{|c}{\textbf{CS (\%)}}  \\ \cline{2-13}  
        ~ & FT & ZO & BO & FT & ZO & BO & FT & ZO & BO & FT & ZO & BO  \\ \hline
        SOTA Model & 58.7 & / & / & 7.4 &  / & / & 25.7 & / & / & 24.0 &  / & /   \\ \hline 
        \codegen{}-6B  & 46.3 & -46.0 & -2.7 & 70.4 & -69.0 & -47.0 & 43.2 & -43.2 & +13.0 & 19.0 & -19.0 & -6.0   \\ \hline \hline
        \chatglm{}-6B &  \textbf{\underline{61.0}} & -53.9 & -6.8 & \textbf{\underline{96.9}} & -79.4 & -84.1 & 55.7 & -54.0 & -9.5 & 44.0 & +1.0 & +10.0  \\ \hline
        \vicuna{}-7B &  54.6 & -0.6 & -0.5 & 77.0 & -63.8 & -77.0 & 39.8 & -29.7 & -8.6 & 45.0 & +3.0 & -8.0 \\ \hline
        \alpaca{}-7B & 50.2 & -4.4 & +5.2 & 96.6 & -74.5 & -96.6 & 54.5 & -49.2 & -13.1 & \textbf{\underline{67.0}} & -35.0 & -6.0  \\ \hline
        \dolly{}-7B &  53.1 & -20.0 & -3.2 & 95.7 & -74.4 & -72.2 & 52.0 & -50.1 & -1.0 & 57.0 & -45.0 & -43.0  \\ \hline
        \stablelm{}-7B & 47.2 & -2.9 & -3.8 & 69.9 & -45.6 & -69.9 & 5.6 & -4.5 & +38.8 & 2.0 & +28.0 & +17.0  \\ \hline
        \codealpaca{}-7B & 55.1 & -3.2 & -4.8 & 95.6 & -94.2 & -85.3 & \textbf{\underline{57.8}} & -53.4 & -22.7 & 58.0 & -49.0 & -24.0  \\ \hline
        
	\end{tabular}
    \end{threeparttable}
	\end{adjustbox}
\end{table*}

\subsection{Fine-tuned Performance (RQ3)}
In RQ3, we investigate the performance of instruction-tuned LLMs after being fine-tuned on each task. Table~\ref{table:tunedResult} presents the fine-tuned performance of instructed LLMs on each task, where ``FT'' column presents the fine-tuned performance while ``ZO'' and ``BO'' columns present the one-shot performance and one-shot performance against the fine-tuned performance. We also include the performance of fine-tuning the SOTA small models on each task (``SOTA model'' Row). 

\parabf{Fine-tuning v.s. Zero/One-shot.} Fine-tuning the instructed LLMs  on specific downstream tasks further improves the model performance compared to the zero-shot/one-shot setting. As shown in the table, the best performance in each task is achieved by fine-tuning the instructed LLM on the specific task. Previous work~\cite{mosbach2023fewshot, DBLP:conf/nips/LiuTMMHBR22, sanh2022multitask} also reveals similar observations that parameter-efficient fine-tuning can outperform in-context learning (\ie{} few-shot learning) on models such as GPT-3, which are consistent with our results; but our evaluation focuses on more recent instruction-tuned LLMs in the code comprehension and generation domain.

\begin{table}[htb]
	\centering
	\small
	\caption{\llm{} trainable parameters}\label{table:modelsParam}
	
	\begin{adjustbox}{width=0.97\columnwidth}
	   	
	\begin{tabular}{c|c|c}
		\hline
        \textbf{Group} & \textbf{LLM}  &  \textbf{Training Parameter}  \\ \hline 
\multirow{6}{*}{\tabincell{c}{Instruction-tuned \\ LLM with \lora{}}} & ChatGLM-6B           & 3M   \\
                                        & Vicuna-7B            & 8M   \\
                                        & Alpaca-7B            & 8M   \\
                                        & Dolly-7B             & 8M   \\
                                        & StableLM-7B          & 6M   \\
                                        & CodeAplaca-7B        & 8M   \\ \hline
\multirow{4}{*}{SOTA model}         
                                        & CoTexT-220M          & 220M \\
                                        & CodeGPT-adapted-120M & 120M \\
                                        & CodeT5-220M          & 220M \\
                                        & PLBART-140M          & 140M \\
        
        \hline
	\end{tabular}
	\end{adjustbox}
\end{table}

\parabf{Instructed LLMs v.s. SOTA models.} After being fine-tuned on the same training dataset of each task, most instruction-tuned LLMs outperform the small SOTA models. As mentioned in Section~\ref{sec:setup:procedure}, for the resource limits, we fine-tune each instruction-tuned LLM with the parameter-efficient tuning strategy \lora{}, thus only a small number of parameters would be updated during fine-tuning; for small SOTA models, in line with previous work~\cite{niu2023empirical}, all of their parameters would be updated during fine-tuning. Table~\ref{table:modelsParam} presents the number of parameters updated in fine-tuning for each model. Interestingly, the number of updated parameters in instructed LLMs is actually much smaller than the SOTA small models (\eg{} < 10M v.s. > 100M); however, the instruction-tuned LLMs exhibit much higher performance than SOTA models after fine-tuning. 
The reason might be that the original instructed LLM is already quite powerful (as confirmed by our RQ1) and fine-tuning a small number of parameters is sufficient for further improving the model performance on the downstream task. 
In addition to the small SOTA models, the large model without instruction tuning (\ie{} \codegen{}-6B) still performs worse than instruction-tuned models after both being fine-tuned, indicating that the benefits from instruction tuning cannot be easily mitigated in the subsequent fine-tuning phase.

\finding{Fine-tuning the instructed LLMs can further improve the model performance on downstream code comprehension and generation tasks compared to the zero-shot/one-shot performance. Besides, after fine-tuning on the same downstream task dataset, instructed  LLMs outperform both the small SOTA models and similar-scaled LLMs without instruction tuning, indicating the large benefits of instruction tuning.}

\begin{figure}[htb]
    \centering
    \includegraphics[width=0.5\textwidth]{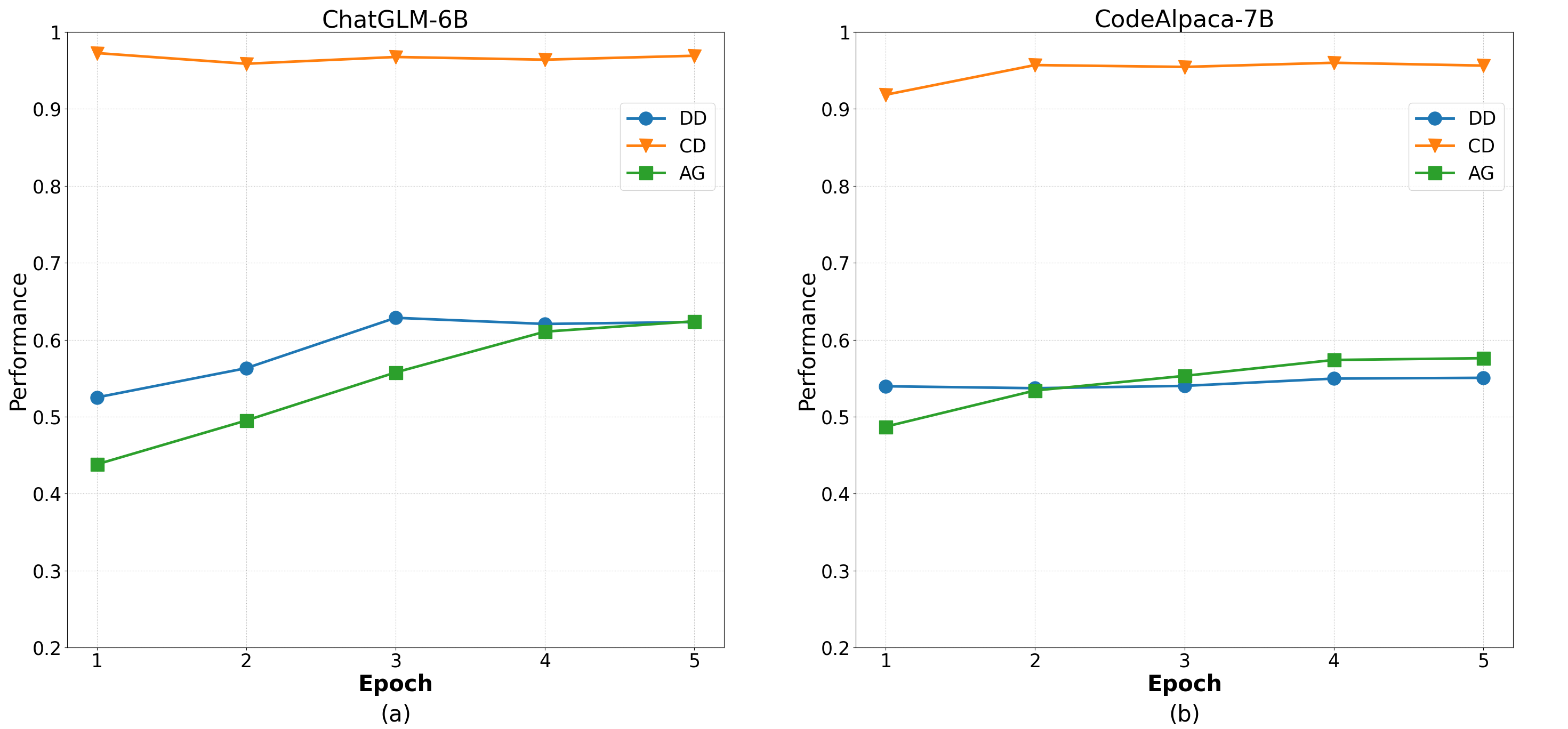}
    \caption{Performance with increasing fine-tuning epochs}
    \label{figure:epochInfluence}
\end{figure}

\parabf{Fine-tuning Epoch.} The results above are based on fine-tuning each model for five epochs. We then present the performance trend of different fine-tuning epochs. For space limits, we present the results of two instructed  LLMs, \ie{} general-domain \chatglm{}-6B and code-specific \codealpaca{}-7B in Figure~\ref{figure:epochInfluence}. As shown in the figure, the model performance reaches a plateau after three epochs, indicating that the instruction-tuned LLMs can quickly adapt to downstream tasks with a few number of fine-tuning epochs. 

\finding{Instruction-tuned LLMs can quickly adapt to downstream code comprehension and generation tasks by being fine-tuned with a few numbers (\eg{} three) of epochs.}

\vspace{-3mm}
\begin{table*}[htb]
	\centering
	\small
	\caption{Time and memory cost of the \llm{} for tuning and inference in each task, evaluated on a single A800-80G GPU.}  
  \label{table:cost}

	\begin{adjustbox}{width=1.9\columnwidth}
	\begin{tabular}{c|r|r|r|r|r|r|r|r}
		\hline
  
        \multirow{2}{*}{\textbf{Model}} & \multicolumn{2}{c|}{\textbf{DD Task}} & \multicolumn{2}{c|}{\textbf{CD Task}} & \multicolumn{2}{c|}{\textbf{AG Task}} & \multicolumn{2}{c}{\textbf{CS Task}} \\ 
        \cline{2-9}
        
        ~ & \textbf{Tuning} & \textbf{Inference} & \textbf{Tuning} &\textbf{Inference} & \textbf{Tuning} & \textbf{Inference} & \textbf{Tuning} & \textbf{Inference}\\ \hline

        {SOTA Model} & 2.5 h (15 G)  & 9.0 s (17 G)& 3.5 h (7 G) &1.0 s (4 G) & 10.0 h (11 G) & 4.5 s (12 G) & 13.0 h (43 G) & 9.0 s (39 G) \\

        {\codegen{}-6B} & 6.0 h (80 G)  & 7.5 m (35 G)& 32.0 h (80 G)  & 9.0 m (35 G)& 28.0 h (77 G) & 7.5 m (35 G) & 16.0 h (77 G) & 8.0 m (35 G)\\
        
        {\chatglm{}-6B} & 7.5 h (48 G)  & 4.5 m (28 G)& 34.0 h (39 G)& 4.5 m (28 G)& 35.0 h (37 G)& 4.5 m (28 G)& 24.0 h (34 G)  & 5.0 m (28 G)\\

        {\vicuna{}-7B} & 5.0 h (80 G)  & 6.0 m (24 G)& 25.0 h (79 G) & 4.5 m (24 G)& 27.0 h (77 G) & 4.5 m (24 G)& 15.0 h (77 G) & 5.0 m (24 G)\\
        
        {\alpaca{}-7B} & 15.5 h (75 G)  & 6.0 m (24 G)& 60.0 h (80 G) & 4.5 m (24 G)& 67.0 h (77 G) & 6.0 m (24 G)& 36.0 h (75 G)  & 5.0 m (24 G)\\

        {\dolly{}-v2-7B} & 12.0 h (80 G)  & 6.0 m (29 G)& 54.0 h (72 G) & 7.5 m (29 G)& 54.0 h (77 G) & 6.0 m (29 G)& 30.0 h (79 G) & 6.0 m (29 G)  \\
        
        {\stablelm{}-7B} & 5.0 h (80 G) &3.0 m (31 G)& 23.0 h (71 G) & 4.5 m (31 G)& 22.0 h (76 G) & 4.5 m (31 G)& 13.0 h (72 G) &4.0 m (31 G)\\
    
        {\codealpaca{}-7B} & 16.0 h (75 G) & 4.5 m (24 G)& 70.0 h (77 G)& 4.5 m (24 G) & 75.0 h (73 G)  & 4.5 m (24 G)& 52.0 h (79 G) & 4.0 m (24 G) \\

        {\dolly{}-v2-12B} & -  & 7.5 m (44 G)& - & 7.5 m (44 G)& - & 7.5 m (44 G)& - & 8.0 m (44 G)\\
        
        {\vicuna{}-13B} & -  & 6.0 m (49 G)& - & 6.0 m (49 G)& - & 10.5 m (49 G)& - & 6.0 m (49 G) \\
        
        {\wizardcoder{}-15B} & -  & 3.0 m (38 G)& - & 4.5 m (38 G)& - & 3.0 m (38 G)& - & 5.0 m (38 G)\\
        
        {\instructcodegen{}-16B} &  -  & 10.5 m (71 G)& - & 10.5 m (71 G)& - & 9.0 m (71 G)& - & 7.0 m (71 G)\\
        
        \hline
	\end{tabular}
	\end{adjustbox}

\end{table*}

\subsection{Costs (RQ4)}~\label{sec:costs}
In RQ4, we analyze both the memory costs and time costs of applying instruction-tuned LLMs. Table~\ref{table:cost} presents the costs of fine-tuning the model per epoch on each code-relevant task (``Tuning'' Column) and the costs during zero-shot inference for 100 testing data items (``Inference'' Column). In addition, we also present the costs of small SOTA models for comparison. 

First, we find that even similar-scaled models take different memory and time costs during fine-tuning and inference. Notably, \chatglm{}-6B is often the most memory-efficient model during fine-tuning. For time costs, \vicuna{}-7B and \stablelm{}-7B are the Top-2 time-efficient models that often take much less fine-tuning time than other instruction-tuned LLMs. Second, the larger models do not necessarily take more time for fine-tuning or inference. For example, larger models such as \wizardcoder{}-15B and \vicuna{}-13B do not necessarily take more inference time than smaller models such as \dolly{}-7B. Lastly, compared to small SOTA models, instructed LLMs take much more memory and time resources, indicating that the small SOTA models might still be the preferred option for the scenario that is sensitive to online latency or limited resources.

\finding{Similar-scaled instructed LLMs vary in memory and time costs, and larger instructed LLMs do not take more time for fine-tuning or inference. Nevertheless, the time and memory cost difference between instructed LLMs and the small SOTA models are non-negligibly huge.}

\section{Implications}
We then summarize main implications based on our findings above. 
\textbf{Instruction-tuned LLMs are very promising on code comprehension and generation.}
Compared to small pre-trained models or large models without instruction, instructed LLMs achieve better performance in zero-shot, few-shot, and even fine-tuning scenarios for code comprehension and generation. 
Thus, instructed LLMs can be preferred in solving code-related tasks if resources allow. While the main purpose of this study is not to deliver a particular ranking of existing instructed LLMs, we recommend \wizardcoder{}, \vicuna{}, and \chatglm{} for researchers and developers who are interested in applying instructed LLM on code-related tasks, given their overall stable performance on our studied tasks and settings. 

\textbf{Fine-tuning is mostly the silver bullet to get the best performance of the instructed LLMs on code comprehension and generation if resources and data allow.} Although instructed LLMs can achieve acceptable performance in both zero-shot and few-shot settings, models might sometimes exhibit unstable or abnormal behaviors during in-context learning; and fine-tuning them specifically on each code-related task could consistently improve the performance, in a rather stable way. 
As shown in our results, with the parameter-efficient tuning strategy \lora{}, the instructed models can adapt to the downstream task with just a few fine-tuning epochs.
Therefore, we recommend efficiently fine-tuning the instructed LLMs if there are available resources and datasets. 

\textbf{Recommended shot-selection strategies in the few-shot setting.} Although at most cases adding demonstration examples could further improve the model capability on code comprehension and generation tasks, the improvements can be unstable or even negative especially for classification problems or with long input prompts. Therefore, for generation problems, we recommend providing instructed LLMs with similar demonstration examples (\eg{} BM25-based shot  selection strategy); for classification problems, there is no dominantly-better shot selection strategy, and in particular, we suggest balancing the input length and the inclusion of demonstration examples since adding examples might have limited help but elongate the input to decrease the model performance.

\textbf{Trade-offs between performance and costs.} Although having better performance on code generation and comprehension tasks, instructed LLMs still take non-negligibly more time in  tuning and inference than small SOTA models. Therefore, for the cases strictly requiring immediate online response or with limited tuning resources, fine-tuned small SOTA models might still be the safe choice to trade off between reasonable performance and costs. In addition, among instructed LLMs, we find larger models do not necessarily take longer time for inference; thus moderately speaking, we recommend \wizardcoder{}, \vicuna{}, and \chatglm{} as reasonable options with better trade-offs between performance and costs. 

\textbf{Future directions.} We further  suggest the following future directions on instructed LLMs for code comprehension and generation. First, we suggest future work on more sound and thorough evaluation approaches for instructed LLMs on code-related tasks, including implementing automatic evaluation protocol, constructing practical code-related benchmarks, and designing sound evaluation metrics, which could benefit the field of evaluating the large body of instructed LLMs systematically. Second, we suggest future work on more efficient tuning and inference of instructed LLMs. While instructed LLMs show promising and powerful performance on code comprehension and generation, they still take much more memory and time resources than small pre-trained models. Thus, building more efficient tuning and inference approaches would definitely put the instructed LLMs towards more practical usage in the code-related domain. Lastly, we suggest more efforts on building instructed LLMs oriented to code comprehension and generation tasks. Given the high diversity in software maintenance and development activities, there are actually a great amount of code-related instructions that could be used to tune the LLMs, which could help build more universal and powerful instructed LLMs to facilitate software engineering activities. 

\section{THREATS TO VALIDITY}
The validity of our results might be threatened by the following issues. 
First, the findings might be limited by the prompt used in our experiments. To construct a reasonable prompt for studied instruction-tuned LLMs, we mainly follow the common practice in prompt design, such as adopting the default system prompt if provided by each instruction-tuned LLM itself and reusing the previous task instruction used in previous code comprehension and generation tasks as mentioned in Section~\ref{sec:setup:prompt}. In addition, we further perform a small-scale pilot study to manually refine the prompt on a small scale dataset (\ie{} five data items that are not overlapped with our testing dataset) to ensure the prompt does not trigger abnormal behaviors of studied models.
Even so, we still cannot guarantee we incorporate the optimal prompt for each task and our findings might be biased by the prompt used in our experiments. It would be important future work to extensively explore the impact of different prompts. 
Second, our findings might be limited to the tasks and datasets used in our experiments,  and cannot be generalized to other tasks or datasets. To mitigate these issues, we construct our datasets by sampling the well-established datasets that have been used in previous work~\cite{niu2023empirical}. In addition, we select four representative code comprehension and generation tasks that have been widely studied in previous work~\cite{niu2023empirical,DBLP:conf/sigsoft/WangYGP0L22,zeng2022extensive}, covering both classification and generation problems. We plan to include more datasets and tasks in the future work. 
Third, the potential data leakage between instruction dataset and testing dataset might induce biases in our results. To mitigate this issue, we compare the instruction dataset of studied models (if released) and the testing dataset and find there is no overlapping instruction. 
Lastly, the evaluation metrics might also threaten the validity of our results. To mitigate the issue, we adopt widely-used metrics for most tasks, such as accuracy, F1, and exact match metrics. In addition for the task with open responses (\ie{} code summarization), we do not follow the problematic metric BLEU but adopt the recent evaluation approach~\cite{zheng2023judging, gudibande2023false, wang2023large, bubeck2023sparks,chiang2023large,dettmers2023qlora,dubois2023alpacafarm,peng2023instruction} for instruction-tuned LLMs by leveraging the powerful LLM ChatGPT as the judge.

\section{Related Work}

\indent\textbf{Evaluation of instruction-tuned LLMs. }
Given the emerging popularity of instruction-tuned LLMs recently, many efforts~\cite{DBLP:journals/corr/abs-2306-04757,zhao2023survey, ji2023exploring, chung2022scaling} have been dedicated to evaluating the instruction-tuned LLMs. As summarized by the latest survey~\cite{DBLP:journals/corr/abs-2307-03109}, existing evaluations for instructed models have covered a broad scope, including not only general NLP tasks~\cite{raheja2023coedit,chakrabarty2022help} (\eg{} sentiment analysis, text classification, and semantic understanding) but also specific domains~\cite{labrak2023zeroshot} (\eg{} medical, education, and agent applications). However, little evaluation of instructed LLMs is diving into the software engineering domain, except the NL-to-Code task~\cite{chen2021evaluating,50670,touvron2023llama, luo2023wizardcoder} (\ie{} generating a function for the given natural language description), which is only one of the code-related tasks in software development and maintenance. While there is an emerging trend of leveraging the instructed models such as ChatGPT and Codex~\cite{DBLP:conf/icse/LemieuxILS23, DBLP:journals/corr/abs-2307-04346, DBLP:conf/icse/KangYY23,zan2023large} on more software engineering tasks (such as test generation and program repair), these commercial models are closed-source, thus lacking transparency and reproducibility. In this work, we make the first attempt to evaluate a wide spectrum of \textit{open-source} instructed LLMs on four representative code comprehension and generation tasks.

\parabf{Evaluation of pre-trained models on code-related tasks.} Given the recent advance in pre-trained models, researchers have extensively evaluated the pre-trained, fine-tuned and prompted model performance on code-related tasks~\cite{DBLP:conf/icse/TufanoPB23,zeng2022extensive,niu2023empirical, DBLP:conf/sigsoft/WangYGP0L22}. For example, Zeng~\et{}~\cite{zeng2022extensive} evaluate eight pre-trained models on seven code-related tasks; more recently, Niu~\et{}~\cite{niu2023empirical} perform a comprehensive evaluation on 19 pre-trained models across 13 code-related tasks. Wang~\et{}\cite{DBLP:conf/sigsoft/WangYGP0L22} compare the fine-tuned performance and prompt-tuned performance of two pre-train models on three code-related tasks. Existing work mainly focuses on small pre-trained models without instruction tuning (such as CodeT5 and CodeBERT), and our work evaluates 10  open-source instruction-tuned LLMs on code-related tasks for the first time.

\section{conclusion}
In this work, we conduct the first systematic study to evaluate the performance of instruction-tuned \llm{}s on code comprehension and generation tasks. Our experiments include 10 recent open-source instructed LLMs with additional five baseline models on four representative code-related tasks. 
Our key findings are as follows.
First, for the \textit{zero-shot} setting, we find that instructed LLMs are very competitive on code comprehension and generation tasks and sometimes even better than small SOTA models specifically fine-tuned on each downstream task.  
Second, for the \textit{few-shot} setting, we find that adding demonstration examples substantially help instructed LLMs perform better on most code comprehension and generation tasks. Third, for the \textit{fine-tuning} setting, we find that fine-tuning could further improve model performance on downstream code comprehension and generation tasks compared to the zero-shot/one-shot performance. Based on our findings, we further present practical implications on model and usage recommendation, performance and cost trade-offs, and future directions.

\balance
\bibliographystyle{unsrt}
\bibliography{ref}

\begin{thebibliography}{10}

\bibitem{brown2020language}
Tom~B. Brown, Benjamin Mann, Nick Ryder, Melanie Subbiah, Jared Kaplan, and
  et~al.
\newblock Language models are few-shot learners.
\newblock In {\em Advances in Neural Information Processing Systems 33: Annual
  Conference on Neural Information Processing Systems 2020, NeurIPS 2020,
  December 6-12, 2020, virtual}, 2020.

\bibitem{touvron2023llama}
Hugo Touvron, Thibaut Lavril, Gautier Izacard, Xavier Martinet, and et~al.
\newblock Llama: Open and efficient foundation language models.
\newblock {\em CoRR}, abs/2302.13971, 2023.

\bibitem{workshop2023bloom}
Teven~Le Scao, Angela Fan, Christopher Akiki, Ellie Pavlick, and et~al.
\newblock {BLOOM:} {A} 176b-parameter open-access multilingual language model.
\newblock {\em CoRR}, abs/2211.05100, 2022.

\bibitem{chowdhery2022palm}
Aakanksha Chowdhery, Sharan Narang, Jacob Devlin, Maarten Bosma, Gaurav Mishra,
  Adam Roberts, and et~al.
\newblock Palm: Scaling language modeling with pathways.
\newblock {\em CoRR}, abs/2204.02311, 2022.

\bibitem{vicuna2023}
Wei-Lin Chiang, Zhuohan Li, Zi~Lin, Ying Sheng, Zhanghao Wu, Hao Zhang, Lianmin
  Zheng, Siyuan Zhuang, Yonghao Zhuang, Joseph~E. Gonzalez, Ion Stoica, and
  Eric~P. Xing.
\newblock Vicuna: An open-source chatbot impressing gpt-4 with 90\%* chatgpt
  quality, March 2023.

\bibitem{alpaca}
Rohan Taori, Ishaan Gulrajani, Tianyi Zhang, Yann Dubois, Xuechen Li, Carlos
  Guestrin, Percy Liang, and Tatsunori~B. Hashimoto.
\newblock Stanford alpaca: An instruction-following llama model.
\newblock \url{https://github.com/tatsu-lab/stanford_alpaca}, 2023.

\bibitem{codealpaca}
Sahil Chaudhary.
\newblock Code alpaca: An instruction-following llama model for code
  generation.
\newblock \url{https://github.com/sahil280114/codealpaca}, 2023.

\bibitem{sanh2022multitask}
Victor Sanh, Albert Webson, Colin Raffel, Stephen~H. Bach, Lintang Sutawika,
  Zaid Alyafeai, and et~al.
\newblock Multitask prompted training enables zero-shot task generalization.
\newblock In {\em The Tenth International Conference on Learning
  Representations, {ICLR} 2022, Virtual Event, April 25-29, 2022}.
  OpenReview.net, 2022.

\bibitem{ouyang2022training}
Long Ouyang, Jeffrey Wu, Xu~Jiang, Diogo Almeida, and et~al.
\newblock Training language models to follow instructions with human feedback.
\newblock In {\em NeurIPS}, 2022.

\bibitem{wei2022finetuned}
Jason Wei, Maarten Bosma, Vincent~Y. Zhao, Kelvin Guu, Adams~Wei Yu, Brian
  Lester, Nan Du, Andrew~M. Dai, and Quoc~V. Le.
\newblock Finetuned language models are zero-shot learners.
\newblock In {\em The Tenth International Conference on Learning
  Representations, {ICLR} 2022, Virtual Event, April 25-29, 2022}.
  OpenReview.net, 2022.

\bibitem{chatgpt}
{\em ChatGPT}.
\newblock \url{https://openai.com/blog/chatgpt}.

\bibitem{DBLP:journals/corr/abs-2307-03109}
Yupeng Chang, Xu~Wang, Jindong Wang, Yuan Wu, Kaijie Zhu, and et~al.
\newblock A survey on evaluation of large language models.
\newblock {\em CoRR}, abs/2307.03109, 2023.

\bibitem{DBLP:journals/corr/abs-2306-04757}
Yew~Ken Chia, Pengfei Hong, Lidong Bing, and Soujanya Poria.
\newblock {INSTRUCTEVAL:} towards holistic evaluation of instruction-tuned
  large language models.
\newblock {\em CoRR}, abs/2306.04757, 2023.

\bibitem{zhao2023survey}
Wayne~Xin Zhao, Kun Zhou, Junyi Li, Tianyi Tang, Xiaolei Wang, and et~al.
\newblock A survey of large language models.
\newblock {\em CoRR}, abs/2303.18223, 2023.

\bibitem{ji2023exploring}
Yunjie Ji, Yong Deng, Yan Gong, Yiping Peng, Qiang Niu, Lei Zhang, Baochang Ma,
  and Xiangang Li.
\newblock Exploring the impact of instruction data scaling on large language
  models: An empirical study on real-world use cases.
\newblock {\em CoRR}, abs/2303.14742, 2023.

\bibitem{chung2022scaling}
Hyung~Won Chung, Le~Hou, Shayne Longpre, Barret Zoph, Yi~Tay, William Fedus,
  Eric Li, Xuezhi Wang, and et~al.
\newblock Scaling instruction-finetuned language models.
\newblock {\em CoRR}, abs/2210.11416, 2022.

\bibitem{raheja2023coedit}
Vipul Raheja, Dhruv Kumar, Ryan Koo, and Dongyeop Kang.
\newblock Coedit: Text editing by task-specific instruction tuning.
\newblock {\em CoRR}, abs/2305.09857, 2023.

\bibitem{chakrabarty2022help}
Tuhin Chakrabarty, Vishakh Padmakumar, and He~He.
\newblock Help me write a poem - instruction tuning as a vehicle for
  collaborative poetry writing.
\newblock In {\em Proceedings of the 2022 Conference on Empirical Methods in
  Natural Language Processing, {EMNLP} 2022, Abu Dhabi, United Arab Emirates,
  December 7-11, 2022}, pages 6848--6863. Association for Computational
  Linguistics, 2022.

\bibitem{labrak2023zeroshot}
Yanis Labrak, Mickael Rouvier, and Richard Dufour.
\newblock A zero-shot and few-shot study of instruction-finetuned large
  language models applied to clinical and biomedical tasks.
\newblock {\em CoRR}, abs/2307.12114, 2023.

\bibitem{chen2021evaluating}
Mark Chen, Jerry Tworek, Heewoo Jun, Qiming Yuan, Henrique~Pond{\'{e}}
  de~Oliveira~Pinto, and et~al.
\newblock Evaluating large language models trained on code.
\newblock {\em CoRR}, abs/2107.03374, 2021.

\bibitem{50670}
Jacob Austin, Augustus Odena, Maxwell~I. Nye, Maarten Bosma, Henryk
  Michalewski, David Dohan, Ellen Jiang, Carrie~J. Cai, Michael Terry, Quoc~V.
  Le, and Charles Sutton.
\newblock Program synthesis with large language models.
\newblock {\em CoRR}, abs/2108.07732, 2021.

\bibitem{luo2023wizardcoder}
Ziyang Luo, Can Xu, Pu~Zhao, Qingfeng Sun, and et~al.
\newblock Wizardcoder: Empowering code large language models with
  evol-instruct.
\newblock {\em CoRR}, abs/2306.08568, 2023.

\bibitem{DBLP:conf/icse/LemieuxILS23}
Caroline Lemieux, Jeevana~Priya Inala, Shuvendu~K. Lahiri, and Siddhartha Sen.
\newblock Codamosa: Escaping coverage plateaus in test generation with
  pre-trained large language models.
\newblock In {\em 45th {IEEE/ACM} International Conference on Software
  Engineering, {ICSE} 2023, Melbourne, Australia, May 14-20, 2023}, pages
  919--931. {IEEE}, 2023.

\bibitem{DBLP:journals/corr/abs-2307-04346}
Vasudev Vikram, Caroline Lemieux, and Rohan Padhye.
\newblock Can large language models write good property-based tests?
\newblock {\em CoRR}, abs/2307.04346, 2023.

\bibitem{DBLP:conf/icse/KangYY23}
Sungmin Kang, Juyeon Yoon, and Shin Yoo.
\newblock Large language models are few-shot testers: Exploring llm-based
  general bug reproduction.
\newblock In {\em 45th {IEEE/ACM} International Conference on Software
  Engineering, {ICSE} 2023, Melbourne, Australia, May 14-20, 2023}, pages
  2312--2323. {IEEE}, 2023.

\bibitem{zan2023large}
Daoguang Zan, Bei Chen, Fengji Zhang, Dianjie Lu, Bingchao Wu, Bei Guan, Yongji
  Wang, and Jian{-}Guang Lou.
\newblock Large language models meet nl2code: {A} survey.
\newblock In Anna Rogers, Jordan~L. Boyd{-}Graber, and Naoaki Okazaki, editors,
  {\em Proceedings of the 61st Annual Meeting of the Association for
  Computational Linguistics (Volume 1: Long Papers), {ACL} 2023, Toronto,
  Canada, July 9-14, 2023}, pages 7443--7464. Association for Computational
  Linguistics, 2023.

\bibitem{DBLP:conf/icse/TufanoPB23}
Rosalia Tufano, Luca Pascarella, and Gabriele Bavota.
\newblock Automating code-related tasks through transformers: The impact of
  pre-training.
\newblock In {\em 45th {IEEE/ACM} International Conference on Software
  Engineering, {ICSE} 2023, Melbourne, Australia, May 14-20, 2023}, pages
  2425--2437. {IEEE}, 2023.

\bibitem{zeng2022extensive}
Zhengran Zeng, Hanzhuo Tan, Haotian Zhang, Jing Li, Yuqun Zhang, and Lingming
  Zhang.
\newblock An extensive study on pre-trained models for program understanding
  and generation.
\newblock In Sukyoung Ryu and Yannis Smaragdakis, editors, {\em {ISSTA} '22:
  31st {ACM} {SIGSOFT} International Symposium on Software Testing and
  Analysis, Virtual Event, South Korea, July 18 - 22, 2022}, pages 39--51.
  {ACM}, 2022.

\bibitem{niu2023empirical}
Changan Niu, Chuanyi Li, Vincent Ng, Dongxiao Chen, Jidong Ge, and Bin Luo.
\newblock An empirical comparison of pre-trained models of source code.
\newblock In {\em 45th {IEEE/ACM} International Conference on Software
  Engineering, {ICSE} 2023, Melbourne, Australia, May 14-20, 2023}, pages
  2136--2148. {IEEE}, 2023.

\bibitem{DBLP:conf/sigsoft/WangYGP0L22}
Chaozheng Wang, Yuanhang Yang, Cuiyun Gao, Yun Peng, Hongyu Zhang, and
  Michael~R. Lyu.
\newblock No more fine-tuning? an experimental evaluation of prompt tuning in
  code intelligence.
\newblock In Abhik Roychoudhury, Cristian Cadar, and Miryung Kim, editors, {\em
  Proceedings of the 30th {ACM} Joint European Software Engineering Conference
  and Symposium on the Foundations of Software Engineering, {ESEC/FSE} 2022,
  Singapore, Singapore, November 14-18, 2022}, pages 382--394. {ACM}, 2022.

\bibitem{wang2021codet5}
Yue Wang, Weishi Wang, Shafiq~R. Joty, and Steven C.~H. Hoi.
\newblock Codet5: Identifier-aware unified pre-trained encoder-decoder models
  for code understanding and generation.
\newblock In {\em Proceedings of the 2021 Conference on Empirical Methods in
  Natural Language Processing, {EMNLP} 2021, Virtual Event / Punta Cana,
  Dominican Republic, 7-11 November, 2021}, pages 8696--8708. Association for
  Computational Linguistics, 2021.

\bibitem{feng2020codebert}
Zhangyin Feng, Daya Guo, Duyu Tang, and et~al. Nan~Duan.
\newblock Codebert: {A} pre-trained model for programming and natural
  languages.
\newblock In {\em Findings of the Association for Computational Linguistics:
  {EMNLP} 2020, Online Event, 16-20 November 2020}, volume {EMNLP} 2020 of {\em
  Findings of {ACL}}, pages 1536--1547. Association for Computational
  Linguistics, 2020.

\bibitem{biderman2023pythia}
Stella Biderman, Hailey Schoelkopf, Quentin Anthony, Herbie Bradley, Kyle
  O'Brien, and et~al.
\newblock Pythia: {A} suite for analyzing large language models across training
  and scaling.
\newblock {\em CoRR}, abs/2304.01373, 2023.

\bibitem{zeng2022glm130b}
Aohan Zeng, Xiao Liu, Zhengxiao Du, Zihan Wang, and et~al.
\newblock {GLM-130B:} an open bilingual pre-trained model.
\newblock In {\em The Eleventh International Conference on Learning
  Representations, {ICLR} 2023, Kigali, Rwanda, May 1-5, 2023}. OpenReview.net,
  2023.

\bibitem{codegpt}
{\em CodeGPT-adapted}.
\newblock
  \url{https://huggingface.co/microsoft/CodeGPT-small-java-adaptedGPT2}.

\bibitem{phan2021cotext}
Long~N. Phan, Hieu Tran, Daniel Le, Hieu Nguyen, James~T. Anibal, Alec
  Peltekian, and Yanfang Ye.
\newblock Cotext: Multi-task learning with code-text transformer.
\newblock {\em CoRR}, abs/2105.08645, 2021.

\bibitem{ahmad-etal-2021-unified}
Wasi~Uddin Ahmad, Saikat Chakraborty, Baishakhi Ray, and Kai{-}Wei Chang.
\newblock Unified pre-training for program understanding and generation.
\newblock In {\em Proceedings of the 2021 Conference of the North American
  Chapter of the Association for Computational Linguistics: Human Language
  Technologies, {NAACL-HLT} 2021, Online, June 6-11, 2021}, pages 2655--2668.
  Association for Computational Linguistics, 2021.

\bibitem{nijkamp2022codegen}
Erik Nijkamp, Bo~Pang, Hiroaki Hayashi, Lifu Tu, Huan Wang, and et~al.
\newblock Codegen: An open large language model for code with multi-turn
  program synthesis.
\newblock In {\em The Eleventh International Conference on Learning
  Representations, {ICLR} 2023, Kigali, Rwanda, May 1-5, 2023}. OpenReview.net,
  2023.

\bibitem{wei2023chainofthought}
Jason Wei, Xuezhi Wang, Dale Schuurmans, Maarten Bosma, Brian Ichter, Fei Xia,
  Ed~H. Chi, Quoc~V. Le, and Denny Zhou.
\newblock Chain-of-thought prompting elicits reasoning in large language
  models.
\newblock In {\em NeurIPS}, 2022.

\bibitem{shanahan2023talking}
Murray Shanahan.
\newblock Talking about large language models.
\newblock {\em CoRR}, abs/2212.03551, 2022.

\bibitem{hoffmann2022training}
Jordan Hoffmann, Sebastian Borgeaud, Arthur Mensch, Elena Buchatskaya, Trevor
  Cai, Eliza Rutherford, and et~al.
\newblock Training compute-optimal large language models.
\newblock {\em CoRR}, abs/2203.15556, 2022.

\bibitem{taylor2022galactica}
Ross Taylor, Marcin Kardas, Guillem Cucurull, Thomas Scialom, Anthony
  Hartshorn, Elvis Saravia, Andrew Poulton, Viktor Kerkez, and Robert Stojnic.
\newblock Galactica: {A} large language model for science.
\newblock {\em CoRR}, abs/2211.09085, 2022.

\bibitem{houlsby2019parameterefficient}
Neil Houlsby, Andrei Giurgiu, Stanislaw Jastrzebski, Bruna Morrone, Quentin
  de~Laroussilhe, Andrea Gesmundo, Mona Attariyan, and Sylvain Gelly.
\newblock Parameter-efficient transfer learning for {NLP}.
\newblock In {\em Proceedings of the 36th International Conference on Machine
  Learning, {ICML} 2019, 9-15 June 2019, Long Beach, California, {USA}},
  volume~97 of {\em Proceedings of Machine Learning Research}, pages
  2790--2799. {PMLR}, 2019.

\bibitem{liu2022ptuning}
Xiao Liu, Kaixuan Ji, Yicheng Fu, Zhengxiao Du, Zhilin Yang, and Jie Tang.
\newblock P-tuning v2: Prompt tuning can be comparable to fine-tuning
  universally across scales and tasks.
\newblock {\em CoRR}, abs/2110.07602, 2021.

\bibitem{lester2021power}
Brian Lester, Rami Al{-}Rfou, and Noah Constant.
\newblock The power of scale for parameter-efficient prompt tuning.
\newblock In {\em Proceedings of the 2021 Conference on Empirical Methods in
  Natural Language Processing, {EMNLP} 2021, Virtual Event / Punta Cana,
  Dominican Republic, 7-11 November, 2021}, pages 3045--3059. Association for
  Computational Linguistics, 2021.

\bibitem{hu2021lora}
Edward~J. Hu, Yelong Shen, Phillip Wallis, Zeyuan Allen{-}Zhu, and et~al.
\newblock Lora: Low-rank adaptation of large language models.
\newblock In {\em The Tenth International Conference on Learning
  Representations, {ICLR} 2022, Virtual Event, April 25-29, 2022}.
  OpenReview.net, 2022.

\bibitem{niu2022deep}
Changan Niu, Chuanyi Li, Bin Luo, and Vincent Ng.
\newblock Deep learning meets software engineering: {A} survey on pre-trained
  models of source code.
\newblock In Luc~De Raedt, editor, {\em Proceedings of the Thirty-First
  International Joint Conference on Artificial Intelligence, {IJCAI} 2022,
  Vienna, Austria, 23-29 July 2022}, pages 5546--5555. ijcai.org, 2022.

\bibitem{lu2021codexglue}
Shuai Lu, Daya Guo, Shuo Ren, Junjie Huang, and et~al.
\newblock Codexglue: {A} machine learning benchmark dataset for code
  understanding and generation.
\newblock In {\em Proceedings of the Neural Information Processing Systems
  Track on Datasets and Benchmarks 1, NeurIPS Datasets and Benchmarks 2021,
  December 2021, virtual}, 2021.

\bibitem{f1}
{\em F1 Metric}.
\newblock \url{https://www.v7labs.com/blog/f1-score-guide}.

\bibitem{DBLP:conf/sigsoft/RoyFA21}
Devjeet Roy, Sarah Fakhoury, and Venera Arnaoudova.
\newblock Reassessing automatic evaluation metrics for code summarization
  tasks.
\newblock In {\em {ESEC/FSE} '21: 29th {ACM} Joint European Software
  Engineering Conference and Symposium on the Foundations of Software
  Engineering, Athens, Greece, August 23-28, 2021}, pages 1105--1116. {ACM},
  2021.

\bibitem{zheng2023judging}
Lianmin Zheng, Wei{-}Lin Chiang, Ying Sheng, Siyuan Zhuang, Zhanghao Wu, and
  et~al.
\newblock Judging llm-as-a-judge with mt-bench and chatbot arena.
\newblock {\em CoRR}, abs/2306.05685, 2023.

\bibitem{gudibande2023false}
Arnav Gudibande, Eric Wallace, Charlie Snell, Xinyang Geng, Hao Liu, Pieter
  Abbeel, Sergey Levine, and Dawn Song.
\newblock The false promise of imitating proprietary llms.
\newblock {\em CoRR}, abs/2305.15717, 2023.

\bibitem{wang2023large}
Peiyi Wang, Lei Li, Liang Chen, Dawei Zhu, Binghuai Lin, Yunbo Cao, Qi~Liu,
  Tianyu Liu, and Zhifang Sui.
\newblock Large language models are not fair evaluators.
\newblock {\em CoRR}, abs/2305.17926, 2023.

\bibitem{bubeck2023sparks}
S{\'{e}}bastien Bubeck, Varun Chandrasekaran, Ronen Eldan, Johannes Gehrke,
  Eric Horvitz, Ece Kamar, and et~al.
\newblock Sparks of artificial general intelligence: Early experiments with
  {GPT-4}.
\newblock {\em CoRR}, abs/2303.12712, 2023.

\bibitem{chiang2023large}
David~Cheng{-}Han Chiang and Hung{-}yi Lee.
\newblock Can large language models be an alternative to human evaluations?
\newblock In Anna Rogers, Jordan~L. Boyd{-}Graber, and Naoaki Okazaki, editors,
  {\em Proceedings of the 61st Annual Meeting of the Association for
  Computational Linguistics (Volume 1: Long Papers), {ACL} 2023, Toronto,
  Canada, July 9-14, 2023}, pages 15607--15631. Association for Computational
  Linguistics, 2023.

\bibitem{dettmers2023qlora}
Tim Dettmers, Artidoro Pagnoni, Ari Holtzman, and Luke Zettlemoyer.
\newblock Qlora: Efficient finetuning of quantized llms.
\newblock {\em CoRR}, abs/2305.14314, 2023.

\bibitem{dubois2023alpacafarm}
Yann Dubois, Xuechen Li, Rohan Taori, Tianyi Zhang, Ishaan Gulrajani, Jimmy Ba,
  Carlos Guestrin, Percy Liang, and Tatsunori~B. Hashimoto.
\newblock Alpacafarm: {A} simulation framework for methods that learn from
  human feedback.
\newblock {\em CoRR}, abs/2305.14387, 2023.

\bibitem{peng2023instruction}
Baolin Peng, Chunyuan Li, Pengcheng He, Michel Galley, and Jianfeng Gao.
\newblock Instruction tuning with {GPT-4}.
\newblock {\em CoRR}, abs/2304.03277, 2023.

\bibitem{zhou2019devign}
Yaqin Zhou, Shangqing Liu, Jing~Kai Siow, Xiaoning Du, and Yang Liu.
\newblock Devign: Effective vulnerability identification by learning
  comprehensive program semantics via graph neural networks.
\newblock In {\em Advances in Neural Information Processing Systems 32: Annual
  Conference on Neural Information Processing Systems 2019, NeurIPS 2019,
  December 8-14, 2019, Vancouver, BC, Canada}, pages 10197--10207, 2019.

\bibitem{6976121}
Jeffrey Svajlenko, Judith~F. Islam, Iman Keivanloo, Chanchal~Kumar Roy, and
  Mohammad~Mamun Mia.
\newblock Towards a big data curated benchmark of inter-project code clones.
\newblock In {\em 30th {IEEE} International Conference on Software Maintenance
  and Evolution, Victoria, BC, Canada, September 29 - October 3, 2014}, pages
  476--480. {IEEE} Computer Society, 2014.

\bibitem{Watson_2020}
Cody Watson, Michele Tufano, Kevin Moran, Gabriele Bavota, and Denys
  Poshyvanyk.
\newblock On learning meaningful assert statements for unit test cases.
\newblock In {\em {ICSE} '20: 42nd International Conference on Software
  Engineering, Seoul, South Korea, 27 June - 19 July, 2020}, pages 1398--1409.
  {ACM}, 2020.

\bibitem{husain2020codesearchnet}
Hamel Husain, Ho{-}Hsiang Wu, Tiferet Gazit, Miltiadis Allamanis, and Marc
  Brockschmidt.
\newblock Codesearchnet challenge: Evaluating the state of semantic code
  search.
\newblock {\em CoRR}, abs/1909.09436, 2019.

\bibitem{chatglm}
{\em ChatGLM}.
\newblock \url{https://github.com/THUDM/ChatGLM-6B}.

\bibitem{chatglm-corpus}
{\em Chinese and English corpus}.
\newblock \url{https://github.com/THUDM/ChatGLM-6B/blob/main/README_en.md}.

\bibitem{alpaca_data}
{\em Alpaca Data}.
\newblock
  \url{https://github.com/tatsu-lab/stanford_alpaca/blob/main/alpaca_data.json}.

\bibitem{sharegpt}
{\em ShareGPT}.
\newblock \url{https://sharegpt.com/}.

\bibitem{DatabricksBlog2023DollyV2}
Mike Conover, Matt Hayes, Ankit Mathur, Jianwei Xie, Jun Wan, Sam Shah, Ali
  Ghodsi, Patrick Wendell, Matei Zaharia, and Reynold Xin.
\newblock Free dolly: Introducing the world's first truly open
  instruction-tuned llm, 2023.

\bibitem{dolly15k}
{\em databricks-dolly-15k}.
\newblock
  \url{https://huggingface.co/datasets/databricks/databricks-dolly-15k}.

\bibitem{stablelm}
{\em StableLM-tuned-Alpha-7B}.
\newblock \url{https://huggingface.co/stabilityai/stablelm-tuned-alpha-7b}.

\bibitem{stablelm-base}
{\em StableLM-base-Alpha-7B}.
\newblock \url{https://huggingface.co/stabilityai/stablelm-base-alpha-7b}.

\bibitem{stablelm-corpus}
{\em Five conversational dataset}.
\newblock \url{https://huggingface.co/stabilityai/stablelm-tuned-alpha-7b}.

\bibitem{codealpaca_data}
{\em CodeAlpaca Data}.
\newblock
  \url{https://github.com/sahil280114/codealpaca/blob/master/data/code_alpaca_20k.json}.

\bibitem{li2023starcoder}
Raymond Li, Loubna~Ben Allal, Yangtian Zi, Niklas Muennighoff, Denis Kocetkov,
  and et~al.
\newblock Starcoder: may the source be with you!
\newblock {\em CoRR}, abs/2305.06161, 2023.

\bibitem{evol-instruct}
{\em CodeAlpaca with Evol-Instruct}.
\newblock \url{https://github.com/nlpxucan/WizardLM/tree/main/WizardCoder}.

\bibitem{instructcodegen}
{\em Instruct-CodeGen-16B}.
\newblock \url{ https://huggingface.co/sahil2801/instruct-codegen-16B}.

\bibitem{instructcodegencorpus}
{\em Code-related instructions}.
\newblock
  \url{https://huggingface.co/datasets/sahil2801/code_instructions_120k}.

\bibitem{falcon40b}
Ebtesam Almazrouei, Hamza Alobeidli, Abdulaziz Alshamsi, Alessandro Cappelli,
  Ruxandra Cojocaru, Merouane Debbah, Etienne Goffinet, Daniel Heslow, Julien
  Launay, Quentin Malartic, Badreddine Noune, Baptiste Pannier, and Guilherme
  Penedo.
\newblock {Falcon-40B}: an open large language model with state-of-the-art
  performance.
\newblock 2023.

\bibitem{wang2021syncobert}
Xin Wang, Yasheng Wang, Fei Mi, Pingyi Zhou, Yao Wan, Xiao Liu, Li~Li, Hao Wu,
  Jin Liu, and Xin Jiang.
\newblock Syncobert: Syntax-guided multi-modal contrastive pre-training for
  code representation, 2021.

\bibitem{experimentdata}
{\em Available data}.
\newblock
  \url{https://anonymous.4open.science/r/An_Empirical_Study_of_InstructionTuned_LLM/README.md}.

\bibitem{nashid2023retrieval}
Noor Nashid, Mifta Sintaha, and Ali Mesbah.
\newblock Retrieval-based prompt selection for code-related few-shot learning.
\newblock In {\em 45th {IEEE/ACM} International Conference on Software
  Engineering, {ICSE} 2023, Melbourne, Australia, May 14-20, 2023}, pages
  2450--2462. {IEEE}, 2023.

\bibitem{bm25}
Stephen~E. Robertson and Hugo Zaragoza.
\newblock The probabilistic relevance framework: {BM25} and beyond.
\newblock {\em Found. Trends Inf. Retr.}, 3(4):333--389, 2009.

\bibitem{peinl2023evaluation}
Ren{\'{e}} Peinl and Johannes Wirth.
\newblock Evaluation of medium-large language models at zero-shot closed book
  generative question answering.
\newblock {\em CoRR}, abs/2305.11991, 2023.

\bibitem{kim2023aligning}
Sungdong Kim, Sanghwan Bae, Jamin Shin, Soyoung Kang, Dong{-}Hyun Kwak,
  Kang~Min Yoo, and Minjoon Seo.
\newblock Aligning large language models through synthetic feedback.
\newblock {\em CoRR}, abs/2305.13735, 2023.

\bibitem{köpf2023openassistant}
Andreas K{\"{o}}pf, Yannic Kilcher, Dimitri von R{\"{u}}tte, Sotiris
  Anagnostidis, Zhi{-}Rui Tam, Keith Stevens, Abdullah Barhoum, Nguyen~Minh
  Duc, Oliver Stanley, Rich{\'{a}}rd Nagyfi, Shahul ES, Sameer Suri, David
  Glushkov, Arnav Dantuluri, Andrew Maguire, Christoph Schuhmann, Huu Nguyen,
  and Alexander Mattick.
\newblock Openassistant conversations - democratizing large language model
  alignment.
\newblock {\em CoRR}, abs/2304.07327, 2023.

\bibitem{ye2023incontext}
Seonghyeon Ye, Hyeonbin Hwang, Sohee Yang, Hyeongu Yun, Yireun Kim, and Minjoon
  Seo.
\newblock In-context instruction learning.
\newblock {\em CoRR}, abs/2302.14691, 2023.

\bibitem{DBLP:conf/emnlp/MinLHALHZ22}
Sewon Min, Xinxi Lyu, Ari Holtzman, Mikel Artetxe, Mike Lewis, Hannaneh
  Hajishirzi, and Luke Zettlemoyer.
\newblock Rethinking the role of demonstrations: What makes in-context learning
  work?
\newblock In {\em Proceedings of the 2022 Conference on Empirical Methods in
  Natural Language Processing, {EMNLP} 2022, Abu Dhabi, United Arab Emirates,
  December 7-11, 2022}, pages 11048--11064. Association for Computational
  Linguistics, 2022.

\bibitem{DBLP:conf/icml/ZhaoWFK021}
Zihao Zhao, Eric Wallace, Shi Feng, Dan Klein, and Sameer Singh.
\newblock Calibrate before use: Improving few-shot performance of language
  models.
\newblock In {\em Proceedings of the 38th International Conference on Machine
  Learning, {ICML} 2021, 18-24 July 2021, Virtual Event}, volume 139 of {\em
  Proceedings of Machine Learning Research}, pages 12697--12706. {PMLR}, 2021.

\bibitem{DBLP:journals/corr/abs-2307-03172}
Nelson~F. Liu, Kevin Lin, John Hewitt, Ashwin Paranjape, Michele Bevilacqua,
  Fabio Petroni, and Percy Liang.
\newblock Lost in the middle: How language models use long contexts.
\newblock {\em CoRR}, abs/2307.03172, 2023.

\bibitem{DBLP:journals/corr/abs-2307-10558}
Shiyang Li, Jun Yan, Hai Wang, Zheng Tang, Xiang Ren, Vijay Srinivasan, and
  Hongxia Jin.
\newblock Instruction-following evaluation through verbalizer manipulation.
\newblock {\em CoRR}, abs/2307.10558, 2023.

\bibitem{hsu2014paired}
{\em Paired t-test}.
\newblock \url{https://statisticsbyjim.com/hypothesis-testing/paired-t-test/}.

\bibitem{codex}
{\em Codex}.
\newblock \url{https://openai.com/blog/openai-codex}.

\bibitem{mosbach2023fewshot}
Marius Mosbach, Tiago Pimentel, Shauli Ravfogel, Dietrich Klakow, and Yanai
  Elazar.
\newblock Few-shot fine-tuning vs. in-context learning: {A} fair comparison and
  evaluation.
\newblock In {\em Findings of the Association for Computational Linguistics:
  {ACL} 2023, Toronto, Canada, July 9-14, 2023}, pages 12284--12314.
  Association for Computational Linguistics, 2023.

\bibitem{DBLP:conf/nips/LiuTMMHBR22}
Haokun Liu, Derek Tam, Mohammed Muqeeth, Jay Mohta, Tenghao Huang, Mohit
  Bansal, and Colin Raffel.
\newblock Few-shot parameter-efficient fine-tuning is better and cheaper than
  in-context learning.
\newblock In {\em NeurIPS}, 2022.

\end{thebibliography}

\end{document}